\documentclass[a4paper,conference]{IEEEtran}
\IEEEoverridecommandlockouts

\usepackage{pdfpages}

\usepackage{times}
\usepackage{epsfig}
\usepackage{graphicx}

\usepackage{xcolor}

\usepackage{subcaption}

\usepackage{amssymb}

\usepackage{amsmath}

\usepackage{times}

\usepackage{pifont}
\usepackage{graphbox}

\usepackage{epstopdf}

\usepackage{siunitx}

\usepackage{booktabs}

\usepackage{breakcites}

\usepackage{calrsfs}
\DeclareMathAlphabet{\pazocal}{OMS}{zplm}{m}{n}

\begin{document}
%
\title{PROPEL: Probabilistic Parametric Regression Loss for Convolutional Neural Networks}

\author{\IEEEauthorblockN{Muhammad Asad\thanks{This work was undertaken when authors were at City, University of London}}
\IEEEauthorblockA{Imagination Technologies\\
Kings Langley, UK\\
masadcv@gmail.com}
\and
\IEEEauthorblockN{Rilwan Basaru}
\IEEEauthorblockA{Onaria Technologies\\
London, UK\\
remi@onariatech.com}
\and
\IEEEauthorblockN{S M Masudur Rahman Al Arif}
\IEEEauthorblockA{ASML\\
Veldhoven, Netherlands\\
masudur.al.arif@asml.com}
\and
\IEEEauthorblockN{Greg Slabaugh}
\IEEEauthorblockA{City, University of London\\
London, UK
\\ greg.slabaugh@gmail.com}
}

\maketitle

\begin{abstract}
In recent years, Convolutional Neural Networks (CNNs) have enabled significant advancements to the state-of-the-art in computer vision. For classification tasks, CNNs have widely employed probabilistic output and have shown the significance of providing additional confidence for predictions. However, such probabilistic methodologies are not widely applicable for addressing regression problems using CNNs, as regression involves learning unconstrained continuous and, in many cases, multi-variate target variables. We propose a PRObabilistic Parametric rEgression Loss (PROPEL) that facilitates CNNs to learn parameters of probability distributions for addressing probabilistic regression problems. PROPEL is fully differentiable and, hence, can be easily incorporated for end-to-end training of existing CNN regression architectures using existing optimization algorithms. The proposed method is flexible as it enables learning complex unconstrained probabilities while being generalizable to higher dimensional multi-variate regression problems. We utilize a PROPEL-based CNN to address the problem of learning hand and head orientation from uncalibrated color images. Our experimental validation and comparison with existing CNN regression loss functions show that PROPEL improves the accuracy of a CNN by enabling probabilistic regression, while significantly reducing required model parameters by $10 \times$, resulting in improved generalization as compared to the existing state-of-the-art.
\end{abstract}
%
\IEEEpeerreviewmaketitle

\section{Introduction}
Convolutional Neural Networks (CNNs) are enabling major advancements in a range of machine learning problems. For classification tasks, the existing state-of-the-art benefits from probability distributions for target class prediction \cite{he2016deep,krizhevsky2012imagenet,shelhamer2017fully,Simonyan15}. These distributions provide additional confidence information that can be useful to determine the level of uncertainty in a given prediction and help resolve ambiguous model predictions \cite{asad2017spore}. Furthermore, for complex learning problems where dividing a given learning task into smaller subtasks can help, probabilistic models provide a flexible framework for combining the predictions from multiple models \cite{asad2017spore,fanello2014learning}.
\begin{figure}[h!]
	\begin{subfigure}[t]{0.49\textwidth}
		\centering
		\includegraphics[trim={0.45cm 0 0.75cm 0.75cm}, clip, width=0.75\textwidth] {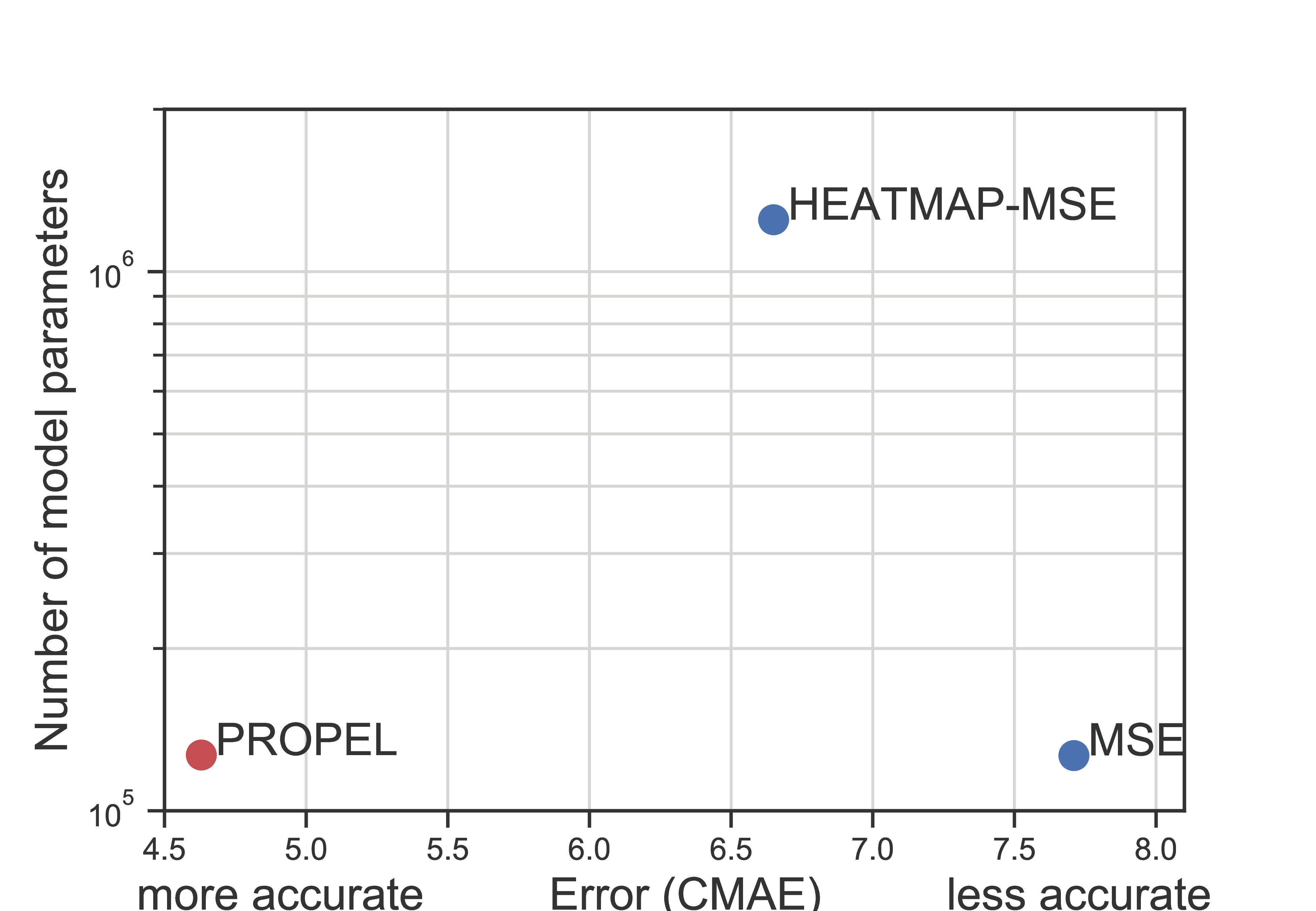}
		\hspace{0.1cm}
		\includegraphics[trim={5cm 0 0 0}, width=0.177\textwidth] {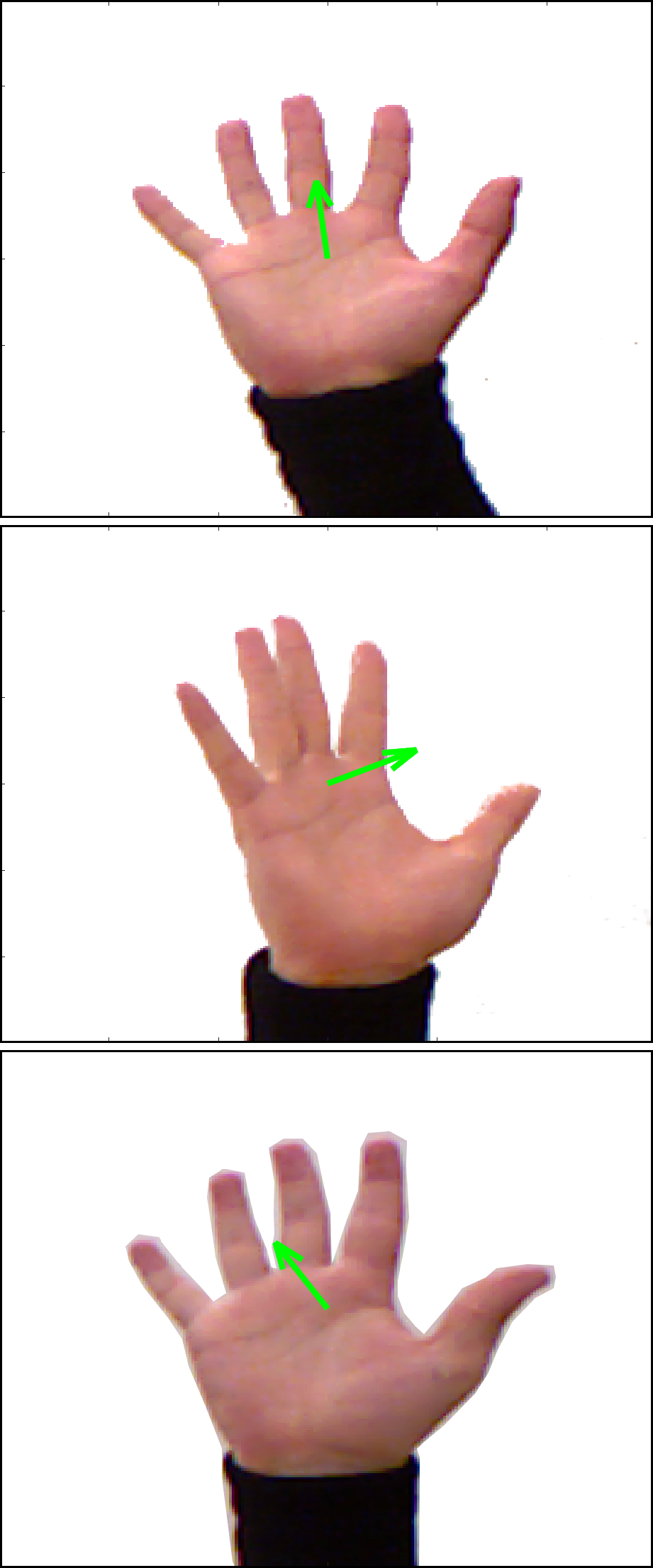}
		\caption{Hand orientation estimation}		
	\end{subfigure}
	\vspace{0.1cm}
	
	\begin{subfigure}[t]{0.49\textwidth}
		\centering
		\includegraphics[trim={0.45cm 0 0.75cm 0.75cm}, clip, width=0.75\textwidth] {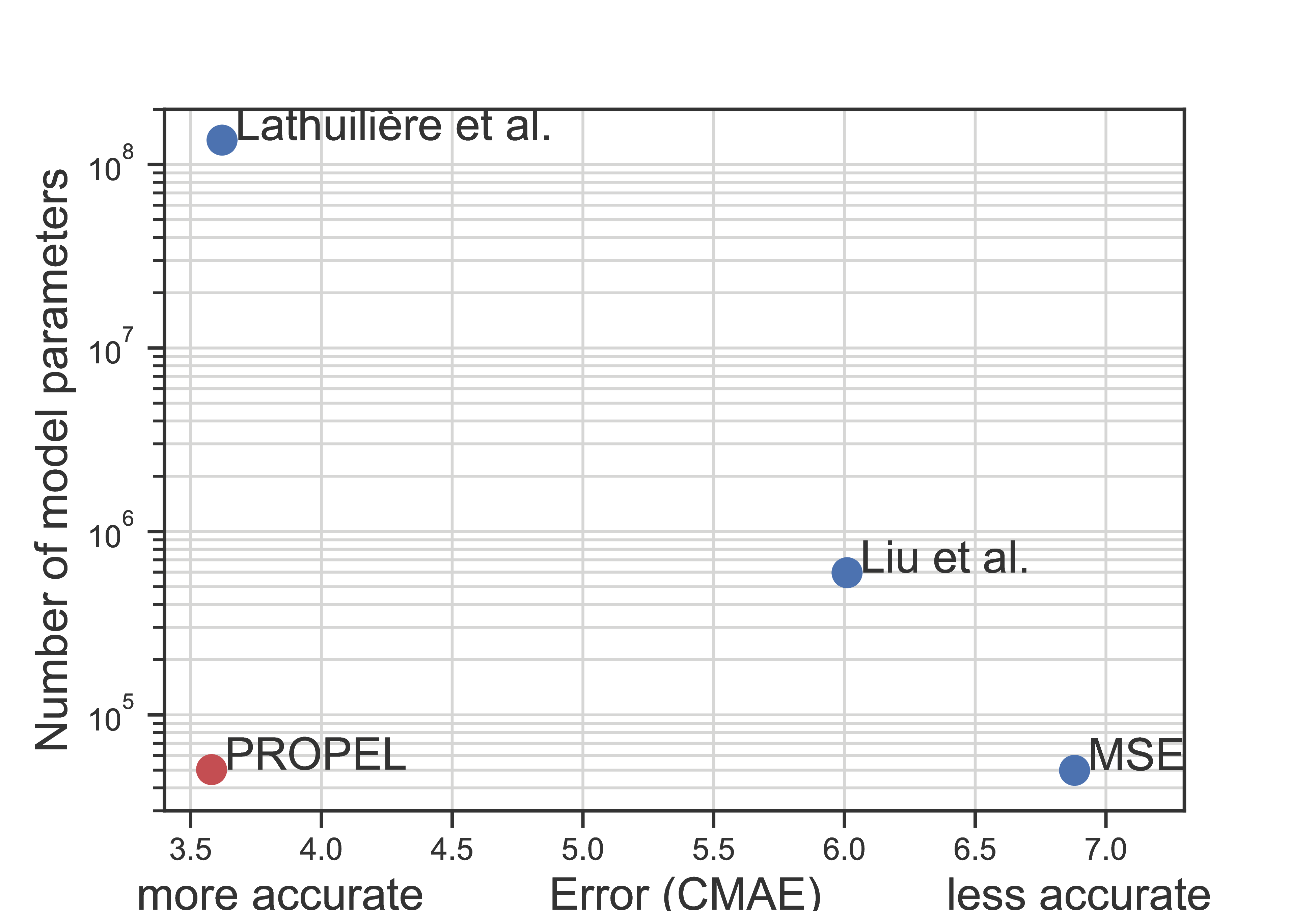}
		\hspace{0.1cm}
		\includegraphics[trim={0.75cm 0 0 0}, width=0.127\textwidth]{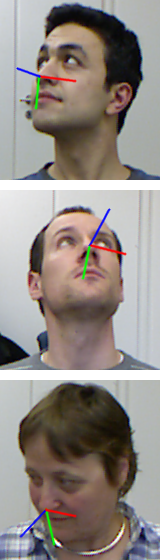}
		\caption{Head orientation estimation}
	\end{subfigure}
	\vspace{0.1cm}

	\caption{Number of model parameters (y-axis, note the log scale) vs accuracy (x-axis) for (a) hand \cite{asad2017spore} and (b) head orientation regression \cite{fanelli2013random}. PROPEL achieves state-of-the-art accuracy for both tasks, with significantly fewer required model parameters as compared to methods Lathuili{\`e}re et al. \cite{lathuiliere2017deep}, Liu et al. \cite{liu20163d}, Mean Squared Error (MSE) and HEATMAP-MSE. \looseness=-1}
	\label{fig:accuracyvsparameters}
\end{figure}

Despite widespread application of CNNs for probabilistic classification, little emphasis has been made for unconstrained probabilistic regression. Regression using a CNN has been restricted to target space learning using Mean Squared Error (MSE) that does not provide a probabilistic output \cite{eigen2014depth,lathuiliere2019comprehensive}. In some existing methods, non-parametric probability heatmaps representing target variables are directly learned using MSE \cite{bulat2016human,ge2016robust,tompson2014real}. 
In order to generate target heatmap distributions, these methods make two assumptions. First, that the target continuous variables can be discretized into heatmap bins without significant loss in accuracy. Second, these variables can be constrained within a specific domain covered by heatmaps. Although necessary, these assumptions limit application of non-parametric probabilistic regression within a specific domain. Moreover, such methods require the output layer of a CNN to represent significantly higher dimensional heatmaps. This contributes to the complexity of the model, requiring larger number of required parameters, increased learning time and training data.\looseness=-1

We propose a novel PRObabilistic Parametric rEgression Loss (PROPEL) that enables a CNN model to learn parametric regression probabilities. Using PROPEL a CNN model can learn parameters of a Mixture of Gaussians distributions in the target space. PROPEL is fully differentiable with an analytic closed-form solution to integrals, allowing it to be used for end-to-end training of existing CNN regression architectures. Moreover, the proposed loss is generalizable for addressing different levels of complexity within prediction probabilities as well as the number of dimensions in multi-variate regression problems. As the output is parameterized by a Mixture of Gaussians, the proposed layer reduces the number of learned parameters as compared to previously used methods based on directly learning probability heatmaps. In this work, we demonstrate PROPEL's ability to enable probabilistic regression by addressing the problem of color image-based hand and head orientation regression \cite{asad2017spore,fanelli2013random}. Both tasks involve learning of ambiguous cases due to symmetries, where multiple orientations can have similar feature representation, which benefit from utilizing probabilistic regression \cite{asad2017spore}. Our experimental results show that PROPEL-based CNN achieve state-of-the-art accuracy, while reducing required model parameters by $10 \times$. These results are summarized in Fig.~\ref{fig:accuracyvsparameters}, showing the accuracy and model parameters trade-off for PROPEL and existing state-of-the-art methods.

The main contributions of this paper are:
\begin{itemize}
	\item We introduce PROPEL which, to the best of our knowledge, is the first fully differentiable loss layer for enabling unconstrained probabilistic parametric regression using CNNs. A novel derivation is provided for the proposed loss using a Mixture of Gaussians distributions;
	\item We present a framework that is generalizable for different number of target dimensions and has the ability to model complex multimodal probabilities with additional Gaussians in the Mixture of Gaussians;
	\item We provide experimental validation and comparisons with existing methodologies and report that the proposed loss outperforms existing state-of-the-art, while reducing the number of required model parameters by $10\times$.
\end{itemize}

\section{Related Work}
\noindent \textbf{Probabilistic Regression using CNN.} Previous work has mostly been focused on exploring non-parametric probabilistic regression for CNNs \cite{alarif2017probabilistic,bulat2016human,ge2016robust,pfister2015flowing,tompson2014real}. These methods first generate the ground truth probability heatmap distributions in the target space. A CNN is then trained using a Mean Squared Error (MSE) loss to learn the mapping of input images onto the heatmaps. In \cite{tompson2014real} Tompson et al. generated and employed single-view 2D heatmaps for hand joints localization using depth images as input. Ge et al. \cite{ge2016robust} extended \cite{tompson2014real} to use multi-view CNN, learning 2D heatmaps for three projections of depth images. Similar methods were also proposed for human pose estimation using a CNN \cite{bulat2016human,pfister2015flowing}. Recent work has also explored the use of 3D heatmaps for hand joint localization \cite{moon2018V2V,payer2019integrating}. Moon et al. \cite{moon2018V2V} proposed to use 3D convolutions to learn mapping of 3D voxelized depth image onto hand joint locations represented as 3D heatmaps. In contrast to using MSE, \cite{alarif2017probabilistic} learns heatmap distributions by utilizing the Bhattacharyya coefficient (BC) as the loss function. Probabilistic interpretation of Euclidean loss has also been previously explored in \cite{pathak2015constrained}.

Lathuili{\`e}re et al. \cite{lathuiliere2017deep} jointly learned both a CNN model, for representation learning, and a Gaussian mixture of linear inverse regressions by utilizing a modified Expectation Maximization (EM) algorithm. Similar to PROPEL, this approach enabled probabilistic regression using CNNs. Crucially, this method required a carefully designed initialization, that included a pre-trained CNN and clustering of data.

Due to computational complexity, non-parametric heatmap distributions only work in problems where the target space can be fully defined within a finite domain, such as the 2D/3D domain used in hand or body joint localization \cite{bulat2016human,ge2016robust,pfister2015flowing,tompson2014real,moon2018V2V,payer2019integrating}. Furthermore, the use of heatmaps require these methods to assume that the continuous target variable can be discretized into heatmap bins without significant accuracy loss. While such assumptions prove useful for introducing and benefiting from the probabilistic regression in CNNs, they limit generalization of these methods and their application to other regression problems. Moreover, the number of model parameters required to directly learn a non-parametric probability heatmap distribution are much higher. The use of heatmaps also limits target variables to lower dimensional problems, e.g. addressing 3D regression problem using 3D heatmaps \cite{moon2018V2V,payer2019integrating} or by decomposing 3D targets into multiple 2D heatmaps \cite{bulat2016human,ge2016robust,tompson2014real}. Additionally, for the joint localization problem, the methods assume that each target joint vary independently, which could have adverse impact on addressing the underlying regression problem \cite{asad2016learning}.\looseness=-1

PROPEL addresses the limitations of existing non-parametric probabilistic regression methods by utilizing a parametric loss function that does not require the target space to be constrained. Furthermore, PROPEL is generalizable in terms of target variable dimension and prediction probability complexity. As indicated by our experimental validation and summarized in Fig.~\ref{fig:accuracyvsparameters}, PROPEL generalizes well when trained on a smaller dataset, while also helping to significantly reduce the number of learned model parameters.
\begin{figure*}[ht!]
	\centering
	\begin{subfigure}[t]{0.16\textwidth}
		\includegraphics[width=\textwidth]{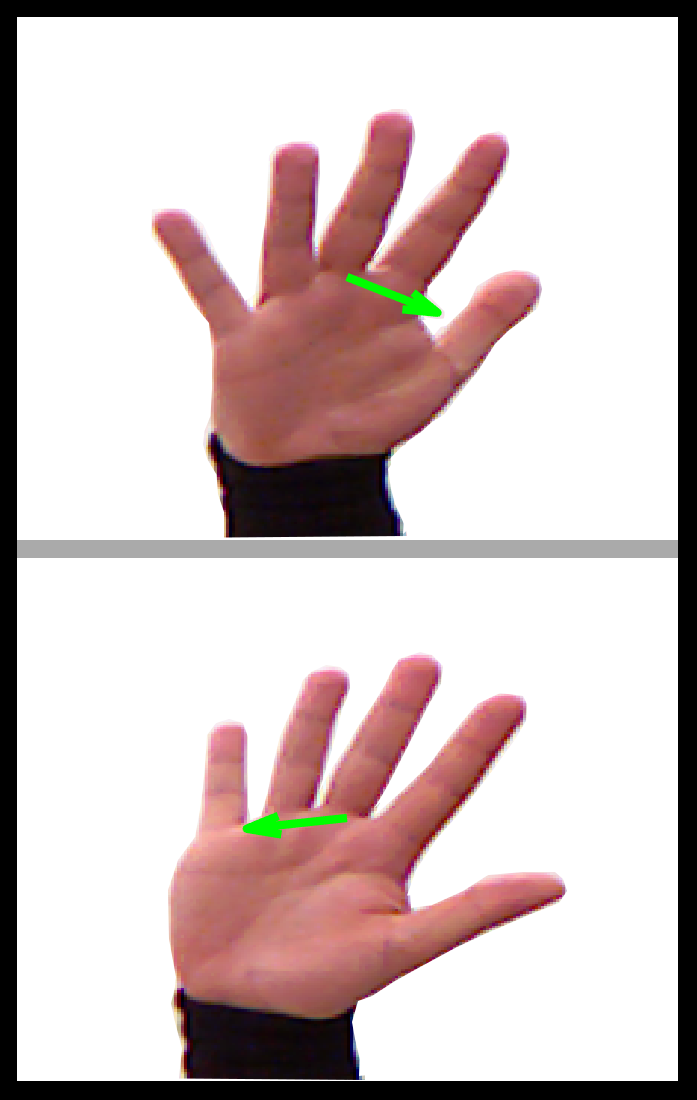}
	\end{subfigure}
	\begin{subfigure}[t]{0.16\textwidth}
		\includegraphics[width=\textwidth]{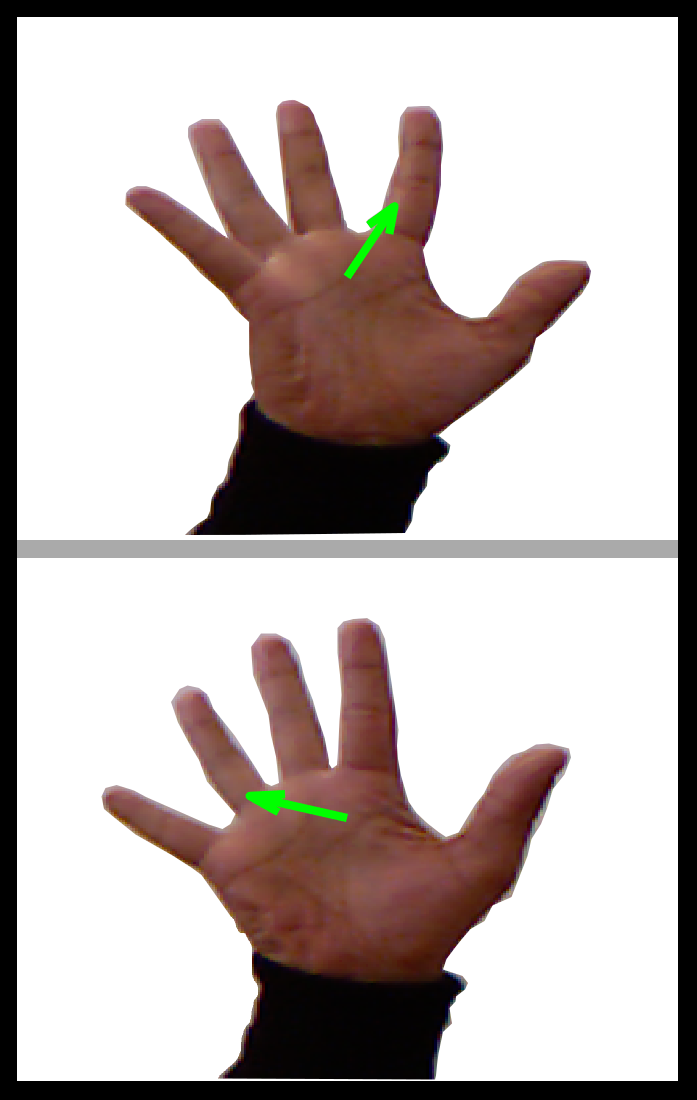}
	\end{subfigure}
	\begin{subfigure}[t]{0.16\textwidth}
		\includegraphics[width=\textwidth]{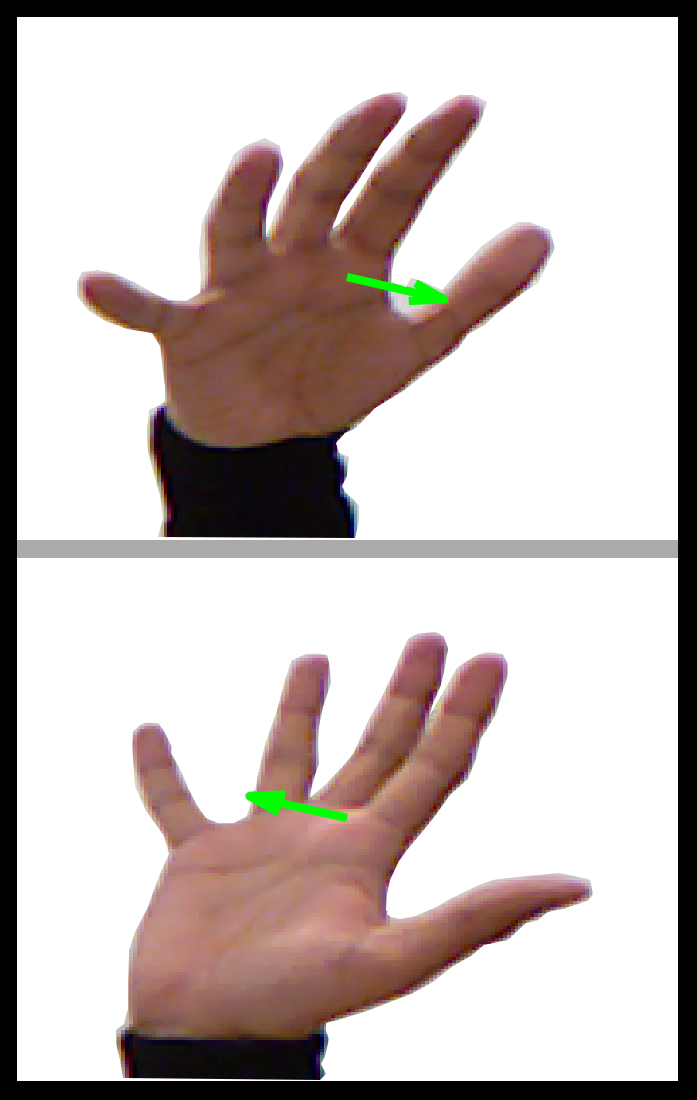}
	\end{subfigure}
	\begin{subfigure}[t]{0.16\textwidth}
		\includegraphics[width=\textwidth]{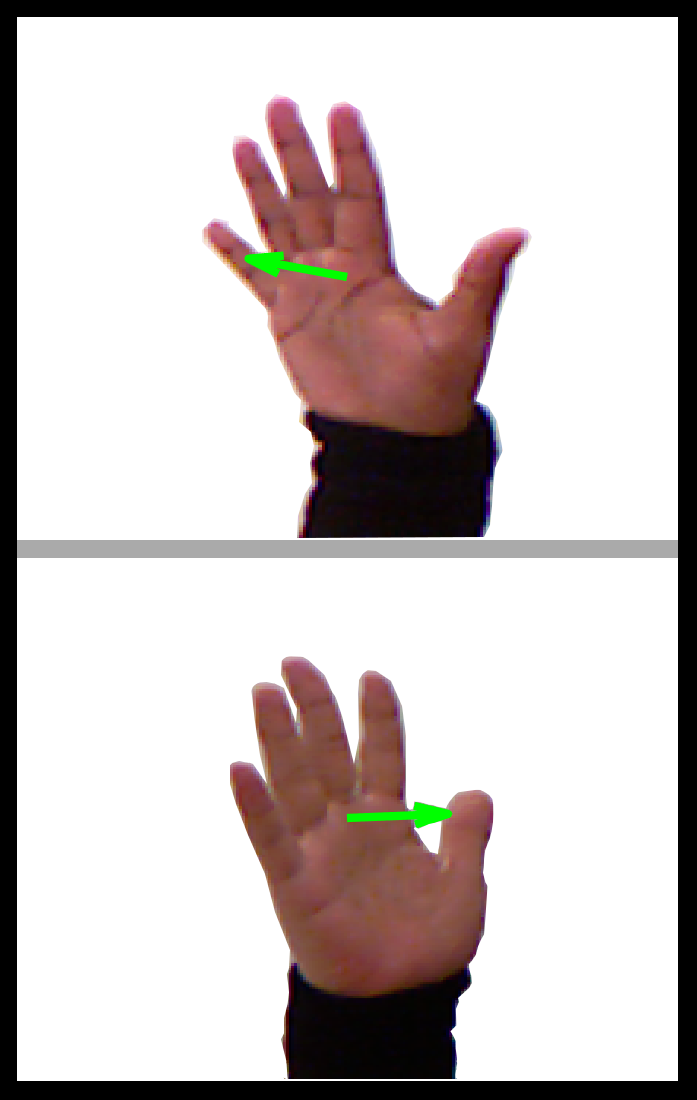}
	\end{subfigure}
	\begin{subfigure}[t]{0.16\textwidth}
		\includegraphics[width=\textwidth]{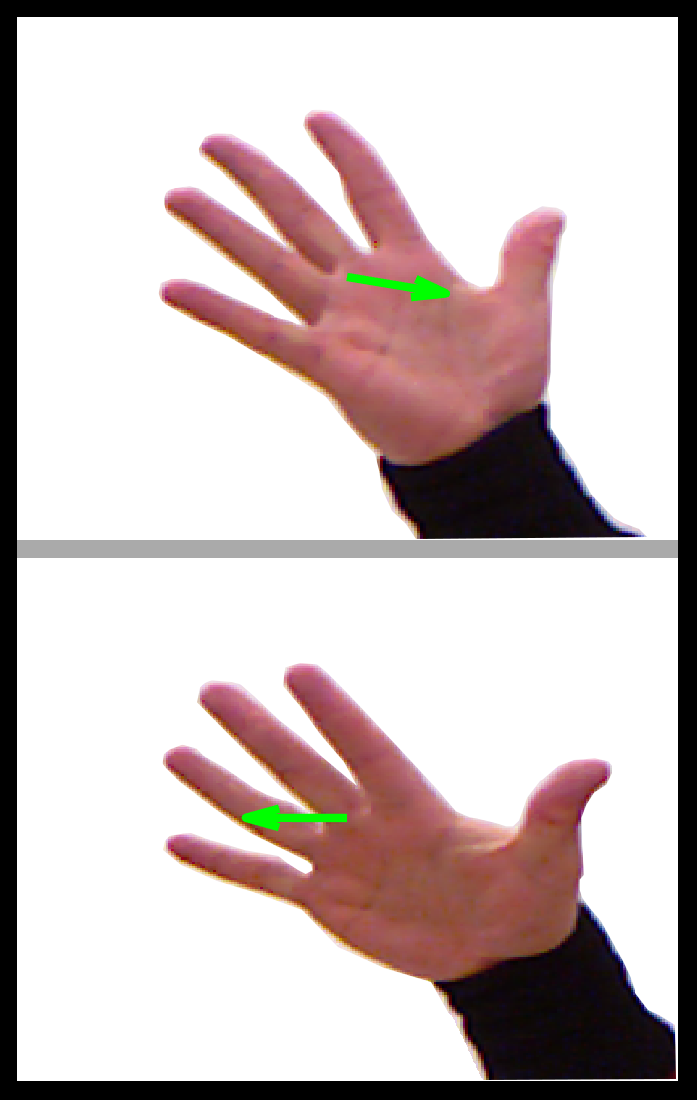}
	\end{subfigure}
	\begin{subfigure}[t]{0.16\textwidth}
		\includegraphics[width=\textwidth]{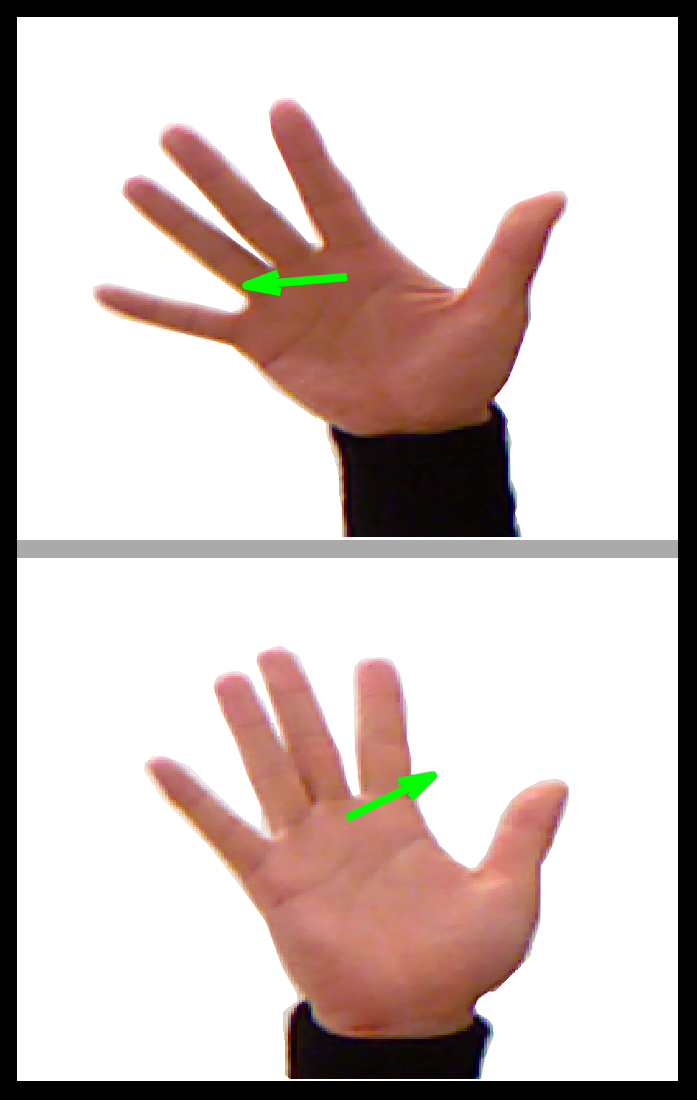}
	\end{subfigure}
	\caption{Symmetry problem due to depth ambiguity in hand orientation. This dataset shows image pairs with different orientations (shown with $\textcolor{green}{\pmb{\uparrow}}$) but similar hand shapes. This motivates the need for multimodal probability distributions in PROPEL.}
	\label{fig:symmetryProblemFigures}
\end{figure*}

\noindent\textbf{Hand orientation Estimation.} Image-based hand orientation regression has only been applied in \cite{asad2014hand,asad2016learning,asad2017spore}. \cite{asad2014hand} utilized two single-variate Random Forest (RF) regressors based on an assumption that the orientation angles vary independently. Later, \cite{asad2016learning} used a multi-layered Random Forest regression method that utilized multi-variate regressors to regress the orientation angles together to exploit target variable dependence. Similarly, \cite{asad2017spore} presented a staged probabilistic regressor, which learned multiple expert Random Forests in stages. Yang et al. \cite{yang2019embedded} proposed a method for localizing hand and inferring only in-plane rotation in images containing multiple hand.

All existing image-based hand orientation regression methods utilize hand-crafted features. In this paper, we show that the task of hand orientation regression can benefit from the representation learning capability of a CNN. Utilizing MSE, we show that a CNN model, without any probabilistic output, can significantly outperform existing methods that use hand-crafted features. We use this model as baseline for our experimental validation and demonstrate that the proposed probabilistic parametric regression loss can outperform existing regression loss functions for CNNs. This is particularly due to PROPEL's ability to handle ambiguity in hand orientation datasets by utilizing probabilistic distributions to comprehensively represent the predictions.

\noindent\textbf{Head orientation Estimation.} A number of existing methods address head orientation regression using color images \cite{drouard2015head,ahn2014real,liu20163d,lathuiliere2017deep}. Drouard et al. \cite{drouard2015head} proposed Gaussian mixture of locally-linear mapping model. This method learned the mapping of HOG-based features onto head orientation, additionally providing probabilistic output. Later, Lathuili{\`e}re et al. \cite{lathuiliere2017deep} improved \cite{drouard2015head} by utilizing a pre-trained CNN for feature extraction. This method, however, relied on a modified EM algorithm for fine-tuning a CNN and learning the mapping of features onto target head orientations. CNN-based head orientation regressors were proposed in \cite{ahn2014real,liu20163d}. Both methods learned using MSE loss, where \cite{liu20163d} utilized an additional synthetic dataset for improving the model. In our experimental validation, we compare against these existing methods and show that, due to the probabilistic reasoning, PROPEL can achieve state-of-the-art accuracy for head orientation estimation, while requiring significantly less learned model parameters.\looseness=-1

\section{Method}
We now describe the details of our proposed PRObabilistic Parametric rEgression Loss (PROPEL). First, we introduce the loss function and the corresponding parametric probabilities. This is followed by the derivation of a novel analytic closed-form solution to the integrals within our loss function that are required for computing the loss over an unconstrained real space. Lastly, we provide details of the partial derivatives of our loss function which allows PROPEL to be integrated with existing CNN architectures using backpropagation.

\subsection{Probabilistic Parametric Regression Loss}
We expand on \cite{sfikas2005analytic} by introducing a regression loss function for measuring loss between ground truth (GT) and predicted parametric probability distributions. The existing measures, such as Bhattacharyya coefficient (BC) and Kullback–-Leibler (KL) divergence, are tractable when distributions are \emph{unimodal}, i.e. each consisting of a single Gaussian. It is well-known that these measures between \emph{multimodal} Mixture of Gaussians have no analytic solution \cite{hershey2007approximating} and at best, one must resort to approximation. However, to address modelling of complex multimodal probabilistic distributions, and handling ambiguous predictions (see Fig~.\ref{fig:PropelVsEuc}), Mixture of Gaussians are essential. This motivates the use of the measure from \cite{sfikas2005analytic} for proposing PROPEL, as this measure has an analytic closed-form solution when distributions are Mixture of Gaussians. We show, for the first time in this paper, how to analytically address neural network regression using a Mixture of Gaussians loss. PROPEL is fully differentiable, enabling us to determine both the loss and gradient of a predicted model distribution $P_{m}$ with respect to a ground truth (GT) distribution $P_{gt}$. Let $\mathbf{x} = \{x_1, x_2, \cdots , x_n\}^\intercal \in \mathbb{R}^n$ define the target prediction space with $n$ dimensions, the proposed loss can be defined for $\mathbf{x}$ as:\looseness=-1

\begingroup
\small
\thinmuskip=\muexpr\thinmuskip*5/8\relax
\medmuskip=\muexpr\medmuskip*5/8\relax 
\begin{align}
L = -\log\left[\frac{2 \int P_{gt} P_{m} \,d\underline{\mathbf{x}} }{\int ({P_{gt}}^2 + {P_{m}}^2) \,d\underline{\mathbf{x}} }\right], \label{eq:mainLossNovel}
\end{align}
\endgroup
where $P_{gt}$ is the GT $n$-dimensional Gaussian probability density function (PDF) defined as:

\begingroup
\small   
\thinmuskip=\muexpr\thinmuskip*5/8\relax
\medmuskip=\muexpr\medmuskip*5/8\relax 
\begin{equation}
P_{gt}^{k} = \frac{e^{-\frac{1}{2}\big[\frac{(x_1-\mu_{x_{1k}})^2}{\sigma_{x_{1k}}} + \cdots + \frac{(x_n-\mu_{x_{nk}})^2}{\sigma_{x_{nk}}} \big]}}{(\sqrt{2\pi})^n\sqrt{\sigma_{x_{1k}} \cdots \sigma_{x_{nk}}}} ,
\label{eq:GTDistribution}
\end{equation}
\endgroup
where $k$ represents a sample selected from a given dataset, $\mu_{x_{1k}}, \mu_{x_{2k}}, \cdots , \mu_{x_{nk}}$ are the GT labels and $\sigma_{\mu_{x_{1k}}}, \sigma_{\mu_{x_{2k}}}, \cdots, \sigma_{\mu_{x_{nk}}}$ are the GT variances associated with $P_{gt}$.

In addition to the GT distribution $P_{gt}$, PROPEL requires a model PDF $P_{m}$, parameters of which are learned by a CNN using PROPEL loss from Equation \ref{eq:mainLossNovel}. The choice of $P_{m}$ determines the ability of a trained model to handle complex variations in target space for a given dataset. Our work is motivated by the limitation of existing regression techniques for addressing ambiguity within a given dataset. Consider the problem of learning hand orientations from uncalibrated color images. As noted in existing literature \cite{asad2016learning}  and shown in Fig.~\ref{fig:symmetryProblemFigures}, the absence of depth information results in ambiguity, where multiple symmetrically opposite hand orientations have similar color images. To address such cases, a unimodal probabilistic distribution proves insufficient (the same is true for probabilistic interpretation of MSE). In contrast, utilizing a multimodal distribution enables the regressor to address such ambiguities by learning to infer multiple hypotheses \cite{asad2017spore,asad2016learning}. In this work, we choose Mixture of Gaussians as the model PDF $P_{m}$ as it facilitates in keeping our derivations simple, while also enabling us to model complex multimodal distributions necessary to address the ambiguity within a dataset. Moreover, as compared to existing state-of-the-art which uses non-parametric heatmaps, the Mixture of Gaussians only require a fraction of parameters (means and variances) to represent the model distribution. This reduces the number of model parameters as well as the overall complexity of a CNN model (see Fig.~\ref{fig:accuracyvsparameters}). Additionally, the number of Gaussians in the Mixture of Gaussians provide flexibility to learn complex probability distributions, resulting in better accuracy while keeping the derivation consistent. For a target space with $n$ dimensions, $P_m$ is defined as:

\begingroup
\small   
\thinmuskip=\muexpr\thinmuskip*5/8\relax
\medmuskip=\muexpr\medmuskip*5/8\relax 
\begin{equation}
P_{m} = \frac{1}{I} \sum_{i=1}^{I} P_i
=\frac{1}{I} \sum_{i=1}^{I} \frac{e^{-\frac{1}{2}\big[\frac{(x_1-\mu_{x_{1i}})^2}{{\sigma_{x_{1i}}}} + \cdots + \frac{(x_n-\mu_{x_{ni}})^2}{{\sigma_{x_{ni}}}} \big]}}{(\sqrt{2\pi})^n\sqrt{\sigma_{x_{1i}}\cdots\sigma_{x_{ni}}}},
\label{eq:modelDistribution}
\end{equation}
\endgroup

\noindent where $P_i$ represents an individual $i^{th}$ Gaussian within a Mixture of Gaussians, $\mu_{x_{1i}} \cdots \mu_{x_{ni}}$ and $\sigma_{x_{1i}} \cdots \sigma_{x_{ni}}$ are model parameters inferred as the output of a CNN network and $I$ is the total number of Gaussians in $P_{m}$. 

Our model distribution does not include covariances since in our experimental validation we found variances to be sufficient for outperforming the state-of-the-art methods. Comparing model distribution $P_{m}$ to a Gaussian Mixture Model (GMM), the Gaussians $P_{i}$ are equally weighted by $\frac{1}{I}$, whereas in a GMM, each Gaussian is given its own weight. However, we note that in our formulation, the Gaussians have different shapes and \emph{heights} resulting from their differing standard deviations. Since the denominator in Equation \ref{eq:modelDistribution} includes standard deviation terms of the form $\sqrt{\sigma_{x_{1i}}\cdots\sigma_{x_{ni}}}$, modes that have less variance (i.e., have more confidence) will have higher peaks in the output. This can be seen in Fig.~\ref{fig:PropelVsEuc} (b), where two modes are ``competing'' for the solution, but the one with less variance has a higher peak and is selected as the solution. By eliminating the weights of a GMM, PROPEL can produce analytic derivatives for back-propagation in deep networks, and at the same time result in giving higher weightage to the peaks in more confident areas of the solution space.

The next section provides a novel analytic closed-form solution of the loss function $L$ such that we can evaluate the integrals over the continuous domain $-\infty$ to $+\infty$.

\subsection{Analytic Solution to Integrals}
\label{sec:AnalyticSolutionMainPaper}
Given the loss function $L$, we can substitute the probability density functions (PDFs) for model $P_{m}$ and GT distribution $P_{gt}$ in Equation \ref{eq:mainLossNovel} and simplify.

\begingroup
\small   
\thinmuskip=\muexpr\thinmuskip*5/8\relax
\medmuskip=\muexpr\medmuskip*5/8\relax  
\begin{align}
&L = -\log\left[\frac{2 \int P_{gt} P_{m} \,d\underline{\mathbf{x}}}{\int({P_{gt}}^2 + {P_{m}}^2) \,d\underline{\mathbf{x}}}\right], \\
&= -\log\left[2 \int P_{gt} P_{m} \,d\underline{\mathbf{x}}\right] + \log \left[\int {P_{gt}}^2 \,d\underline{\mathbf{x}}+ \int {P_{m}}^2 \,d\underline{\mathbf{x}} \right],\\ 
&= -\log\underbrace{\left[ \frac{2}{I} \sum_{i=1}^{I} G(P_{gt}, P_{i}) \right]\rule[-12pt]{0pt}{5pt}}_{\mbox{$T1$}} \nonumber\\
&\,\,\,\,\,\,\,\,+ \log \underbrace{\left[H({P_{gt}}) + \frac{1}{I^2}\sum_{i=1}^{I} H({P_{i}})  + \frac{2}{I^2} \sum_{i < j}^{I} G({P_{i},P_{j}}) \right]\rule[-12pt]{0pt}{5pt}}_{\mbox{$T2$}}, 
\label{eq:LinParts} 
\end{align}
\endgroup 
\noindent where the functions $G(P_{i}, P_{j})$ and $H(P_{i})$ represent the analytic solutions to the integrals $\int {P_{i}P_{j}} \,d\underline{\mathbf{x}}$ and $\int {P_{i}}^2 \,d\underline{\mathbf{x}}$, respectively. $P_i$ and $P_j$ are two multi-variate Gaussian distributions. Both $G(,)$ and $H()$ are defined as follows\footnote{The complete derivation of analytic solution to integrals in functions $G(,)$ and $H()$ is provided in the accompanying supplementary material.}:

\begingroup
\small
\thinmuskip=\muexpr\thinmuskip*5/8\relax
\medmuskip=\muexpr\medmuskip*5/8\relax 
\begin{align}
&G(P_i, P_j) = \int P_{i} P_{j} \,d\underline{\mathbf{x}}, \\
&=  \frac{e^{\big[\frac{2\mu_{x_{1i}}\mu_{x_{1j}} - {\mu_{x_{1i}}}^2 - {\mu_{x_{1j}}}^2}{2(\sigma_{x_{1i}}+\sigma_{x_{1j}})} + \cdots + \frac{2\mu_{x_{ni}}\mu_{x_{nj}} - {\mu_{x_{ni}}}^2 - {\mu_{x_{nj}}}^2}{2(\sigma_{x_{ni}}+\sigma_{x_{nj}})}\big]}}{(\sqrt{2\pi})^n \sqrt{(\sigma_{x_{1i}} + \sigma_{x_{1j}}) \cdots (\sigma_{x_{ni}} + \sigma_{x_{nj}})}}.\label{eq:GTheoremSolved}
\end{align}
\begin{align}
H(P_i) &= \int {P_{i}}^2 \,d\underline{\mathbf{x}} = \frac{1}{(2\sqrt{\pi})^n \sqrt{\sigma_{x_{1i}}\cdots\sigma_{x_{ni}}}}.
\end{align}
\endgroup

Next, we show how the loss $L$ from Equation \ref{eq:LinParts} can be used alongside existing CNN architectures that use backpropagation for training.\looseness=-1

\subsection{Optimization}
\label{sec:Optimization}
In this section we present the partial derivatives of $L$ with respect to model parameters $\mu_{x_{ni}}, \sigma_{x_{ni}}$. These can be used for end-to-end training of a CNN using backpropagation. The partial derivatives of $L$ are:

\begingroup
\scriptsize
\thinmuskip=\muexpr\thinmuskip*5/8\relax
\medmuskip=\muexpr\medmuskip*5/8\relax 
\begin{equation}
\frac{\partial L}{\partial \mu_{x_{ni}}}= -\frac{1}{T1}\left[ \frac{\partial G(P_{gt}, P_{i})}{\partial \mu_{x_{ni}}} \right] + \frac{1}{T2} \left[ \frac{2}{I^2} \sum_{i < j}^{I} \frac{\partial G({P_{i},P_{j}})}{\partial \mu_{x_{ni}}} \right],
\end{equation}
\begin{equation}
\frac{\partial L}{\partial \sigma_{x_{ni}}}= -\frac{1}{T1}\left[ \frac{\partial G(P_{gt}, P_{i})}{\partial \sigma_{x_{ni}}} \right] + \frac{1}{T2} \left[ \frac{1}{I^2} \frac{\partial H({P_{i}})}{\partial \sigma_{x_{ni}}} + \frac{2}{I^2} \sum_{i < j}^{I} \frac{\partial G({P_{i},P_{j}})}{\partial \sigma_{x_{ni}}} \right].
\end{equation}
\endgroup

The partial derivatives $\frac{\partial G(P_i, P_j)}{\partial \mu_{x_{ni}}}$, $\frac{\partial G(P_i, P_j)}{\partial \sigma_{x_{ni}}}$ and $\frac{\partial H(P_i)}{\partial \sigma_{x_{ni}}}$ are:

\begingroup
\small
\thinmuskip=\muexpr\thinmuskip*5/8\relax
\medmuskip=\muexpr\medmuskip*5/8\relax 
\begin{equation}
\frac{\partial G(P_i, P_j)}{\partial \mu_{x_{ni}}} = \frac{(\mu_{x_{nj}} - \mu_{x_{ni}})} {(\sigma_{x_{ni}} + \sigma_{x_{nj}})}\,G(P_i, P_j),
\end{equation}
\endgroup

\begingroup
\small
\thinmuskip=\muexpr\thinmuskip*5/8\relax
\medmuskip=\muexpr\medmuskip*5/8\relax 
\begin{equation}
\frac{\partial G(P_i, P_j)}{\partial \sigma_{x_{ni}}} = [.]\,G(P_i, P_j),\,\,\, \mbox{where},
\end{equation}
\begin{equation}
[.] = \frac{({\mu_{x_{ni}}}^2 + {\mu_{x_{nj}}}^2 - 2\mu_{x_{ni}}\mu_{x_{nj}} - \sigma_{x_{ni}} - \sigma_{x_{nj}})} {2(\sigma_{x_{ni}} + \sigma_{x_{nj}})^2},
\end{equation}
\endgroup

\begingroup
\small
\thinmuskip=\muexpr\thinmuskip*5/8\relax
\medmuskip=\muexpr\medmuskip*5/8\relax 
\begin{equation}
\frac{\partial H(P_i)}{\partial \sigma_{x_{ni}}} = \frac{-1}{2\sigma_{x_{ni}}}\,H(P_i).
\end{equation}
\endgroup

\begin{figure}
	\centering
	\includegraphics[width=1\linewidth, trim={0.5cm, 0.5cm, 0.5cm, 0.5cm}]{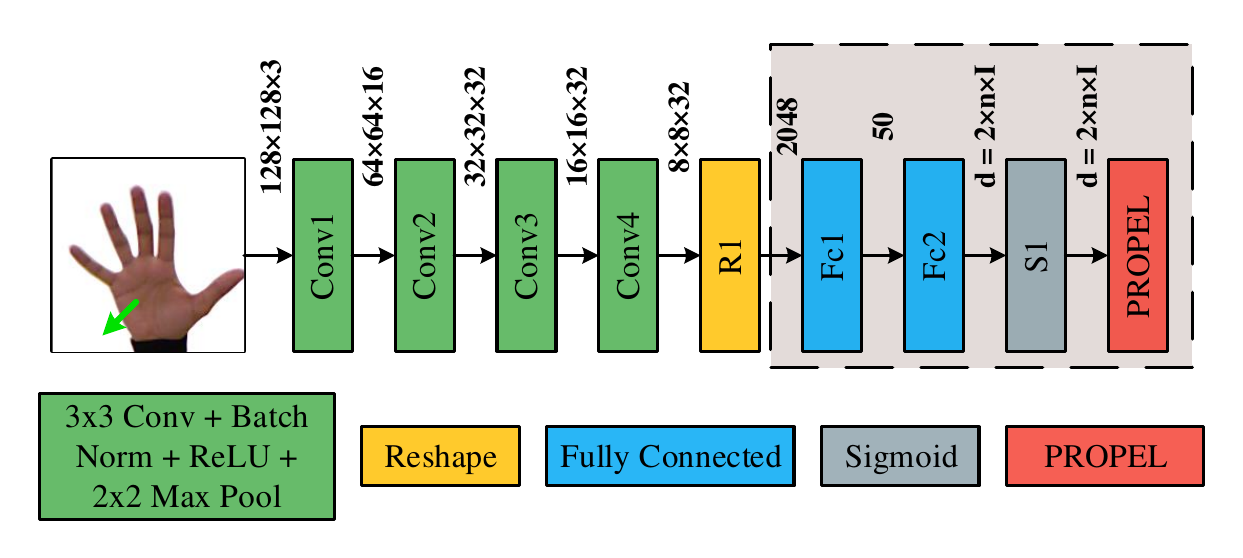}
	\caption{Flowchart showing the CNN regression architecture used for experimental validation of the proposed PROPEL loss (details in Section \ref{sec:networkArchitecture}). The green arrow ($\textcolor{green}{\pmb{\uparrow}}$) shows the target GT hand orientation that is only used during training. Scalar $d$ represents the dimension of the network output. The comparison methods, described in Section \ref{sec:comparisonMethods}, use the same overall architecture, where only the components in the highlighted box are replaced to match output dimensions required by a comparison loss function.}
	\label{fig:CNNArchitecture}
\end{figure}

At each training iteration, PROPEL computes the loss in the forward pass using Equation \ref{eq:LinParts} on the output from the CNN model. For the backward pass, the partial derivatives are used along with the GT labels to backpropagate the error and optimize the parameters using RMSProp \cite{tieleman2012lecture}.

\section{Experimental Validation}
We perform experimental validation of PROPEL by addressing image-based hand and head orientation regression problems. Given a dataset $\pazocal{U} = \{(\mathbf{C}_k, \mathbf{o}_k) \}_{k=1}^{K}$ with $K$ uncalibrated color images $\mathbf{C}_k$ of hands or heads, and the corresponding target orientation vectors $\mathbf{o}_k$, the orientation regression task involve learning the mapping of color images $\mathbf{C}_k$ onto the target orientations $\mathbf{o}_k$. For hands, the orientation vector $\mathbf{o}_k = \{\phi_k, \psi_k\}^\intercal \in \mathbb{R}^2$ contains Azimuth ($\phi_k$) and Elevation ($\psi_k$) angles, resulting from pronation/supination of the forearm and flexion/extension of the wrist, respectively \cite{asad2017spore}. Whereas head orientation is defined by yaw ($\phi_k$), pitch ($\psi_k$) and roll ($\chi_k$) angles, i.e. $\mathbf{o}_k = \{\phi_k, \psi_k, \chi_k\}^\intercal \in \mathbb{R}^3$  \cite{fanelli2013random}. We note that the problem of hand orientation regression is specifically challenging as there may be similar hand shapes that map onto multiple orientations. We evaluate PROPEL on the hand orientation dataset from \cite{asad2017spore}, which contains 9414 images collected from 22 participants. The range of hand orientation angles captured by this dataset are defined within a circular space with $\sqrt{\phi^2 + \psi^2} \le 45 \si{\degree}$. For head orientation, we validate on publicly available BIWI dataset \cite{fanelli2013random}, which contains 10k images from 20 participants. 

\subsection{Network Architecture}
\label{sec:networkArchitecture}
The CNN network utilized for experimental validation of PROPEL is inspired from VGG networks \cite{Simonyan15} (Fig.~\ref{fig:CNNArchitecture}). We keep the number of filters in each layer lower than in \cite{Simonyan15}, as this does not have a significant impact on our model's performance. For hand orientation regression, the input to our network is an RGB color image $\mathbf{C}_k$ of size $128 \times 128 \times 3$, where the method learns to infer parameters for the model distribution $P_m$ in Equation \ref{eq:modelDistribution}. The maximum likelihood estimate (MLE) of the inferred parametric probability distribution is used to get predicted hand orientation angles. Our network has four $3 \times 3$ convolutional layers, where each layer is followed by batch normalization, a rectified non-linear unit (ReLU) and a $2 \times 2$ max pooling layer. The output from the convolutional layers is flattened and fed into two fully connected layers, Fc1 and Fc2 with $50$ and $d = 2 \times n \times I$ neurons, respectively. In our model distribution $P_{m}$, each dimension $n$ requires two parameters, i.e. the mean and the variance of a Gaussian. $P_{m}$ may have $I$ Gaussians resulting in $d = 2 \times n \times I$ parameters as output of Fc2. 

The head orientation validation is done with the same network, however as the size of head images in the BIWI dataset is small, the input images have size $ 64 \times 64$. We also modify Fc2 to enable inference of three-dimensional head orientation, i.e. $n=3$.
\begin{figure}[!tb]
	\centering
	\includegraphics[width=1.0\linewidth]{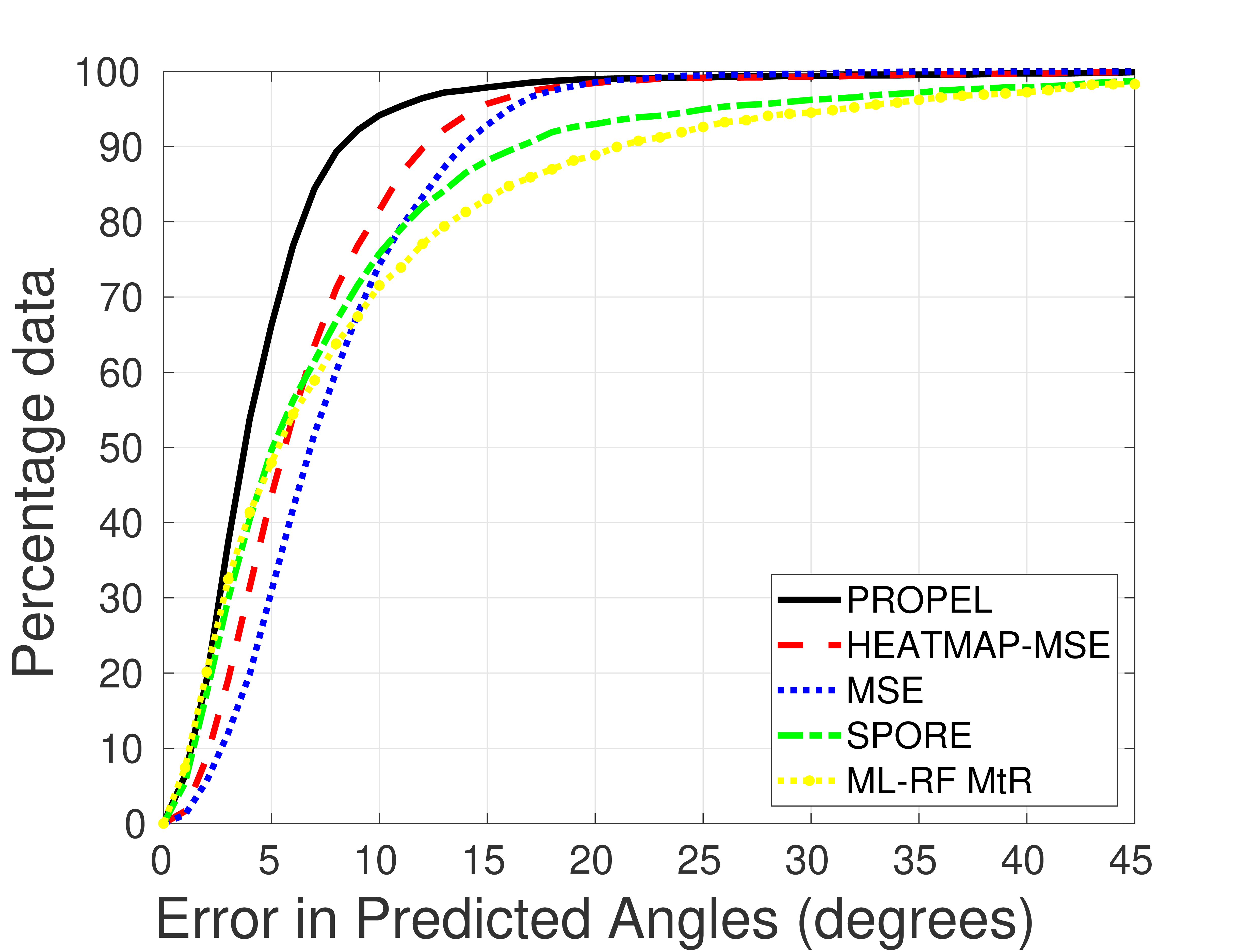}
	\caption{Percentage data versus prediction error shows the percentage data that lies below an error threshold for single-fold validation.}
	\label{fig:PercentageError}
\end{figure}

\subsection{Comparison Methods}
\label{sec:comparisonMethods}
The proposed method is compared with existing state-of-the-art methods for hand orientation regression, namely, Marginalization through Regression (MtR) and Staged PrObabilistic REgression (SPORE) \cite{asad2016learning,asad2017spore}. Both MtR and SPORE use hand-crafted features along with probabilistic Random Forest regressors. Utilizing mean squared error (MSE) loss as baseline, we show that a CNN model, without any probabilistic output, can significantly outperform existing methods that use hand-crafted features. As PROPEL addresses probabilistic regression for a CNN, we compare it with a method inspired from \cite{ge2016robust,tompson2014real} that learns probability heatmap using MSE loss (HEATMAP-MSE). The heatmap distributions are generated by evaluating Equation \ref{eq:GTDistribution} within a $20 \times 20$ grid covering the domain $\phi \in [-45\si{\degree},+45\si{\degree}] ~ \mbox{and} ~ \phi \in [-45\si{\degree},+45\si{\degree}]$. Both MSE and HEATMAP-MSE use the same network architecture as PROPEL, except for the changes in the highlighted box in Fig.~\ref{fig:CNNArchitecture}. MSE uses Fc2 with two neurons, whereas HEATMAP-MSE requires 500 neurons for Fc1 and 400 neurons for Fc2. The additional neurons are required to enable HEATMAP-MSE to learn $20\times20 = 400$ dimensional target heatmap distributions. \looseness=-1
\begin{table}[t!]
	\centering
	\setlength{\tabcolsep}{0.3\tabcolsep}
	\caption{Mean Absolute Error (MAE) in degrees along with number of model parameters for single-fold and 5-fold cross-validation.}
	\resizebox{\columnwidth}{!}{%
		\begin{tabular}{  >{\centering\arraybackslash}m{4.0cm} >{\centering\arraybackslash}m{1.5cm}   >{\centering\arraybackslash}m{1.5cm} >{\centering\arraybackslash}m{1.5cm}
				>{\centering\arraybackslash}m{2.0cm}	}
			\toprule
			\multicolumn{5}{c}{\textbf{Single-fold Validation}} \\
			\midrule
			\textbf{Method} &\textbf{Azimuth ($\phi$)} & \textbf{Elevation ($\phi$)}& \textbf{CMAE} & \textbf{No. of Parameters} \\ \midrule
			PROPEL (proposed) & $\mathbf{5.27}\si{\degree}$ & $\mathbf{3.99}\si{\degree}$ & $\mathbf{4.63}$ & $126,890$\\ 
			HEATMAP-MSE \cite{ge2016robust} & ${7.22}\si{\degree}$ & ${6.07}\si{\degree}$ & $6.65$ & $1,248,932$\\ 
			MSE & ${8.23}\si{\degree}$ & ${7.18}\si{\degree}$ & $7.71$ & $\mathbf{126,584}$ \\ \midrule
			SPORE \cite{asad2017spore} & ${8.49}\si{\degree}$ & ${7.26}\si{\degree}$ & $7.88$ & -\\ 
			MtR \cite{asad2016learning} & ${9.67}\si{\degree}$ & ${7.97}\si{\degree}$ & $8.82$ & -\\ 
			\toprule
			\multicolumn{5}{c}{\textbf{5-fold Cross-validation}} \\
			\midrule
			PROPEL (proposed) & $\mathbf{11.96}\si{\degree}$ & $\mathbf{10.00}\si{\degree}$ & $\mathbf{10.98}$ & $126,890$\\ 
			HEATMAP-MSE \cite{ge2016robust} & ${13.81}\si{\degree}$ & ${10.42}\si{\degree}$ & $12.12$ & $1,248,932$\\ 
			MSE & ${14.16}\si{\degree}$ & ${11.51}\si{\degree}$ & $12.84$ & $\mathbf{126,584}$\\ \midrule
			SPORE \cite{asad2017spore} & ${15.73}\si{\degree}$ & ${12.95}\si{\degree}$ & $14.34$ & - \\ 
			MtR \cite{asad2016learning} & ${16.16}\si{\degree}$ & ${12.83}\si{\degree}$ & $14.50$ & - \\ 
			\bottomrule
	\end{tabular}}
	\label{tb:evaluationSingleFold}
\end{table}

We also compare PROPEL on the image-based head orientation regression task with existing literature. Performance of PROPEL is compared to existing CNN methods
\cite{lathuiliere2017deep,liu20163d} and a hand-crafted feature-based
method \cite{drouard2015head}. \cite{liu20163d} also utilizes synthetic data, however we only include their results from real data for a fair comparison. Comparison with our MSE loss baseline is also made. We exclude HEATMAP-MSE from this comparison as it becomes unwieldy in 3D, and head orientation is represented in 3D.
\begin{figure}[!tb]
	\centering
	\begin{subfigure}{0.5\textwidth}
		\centering
		\includegraphics[width=0.45\linewidth, trim={1.5cm, 0, 0, 0}]{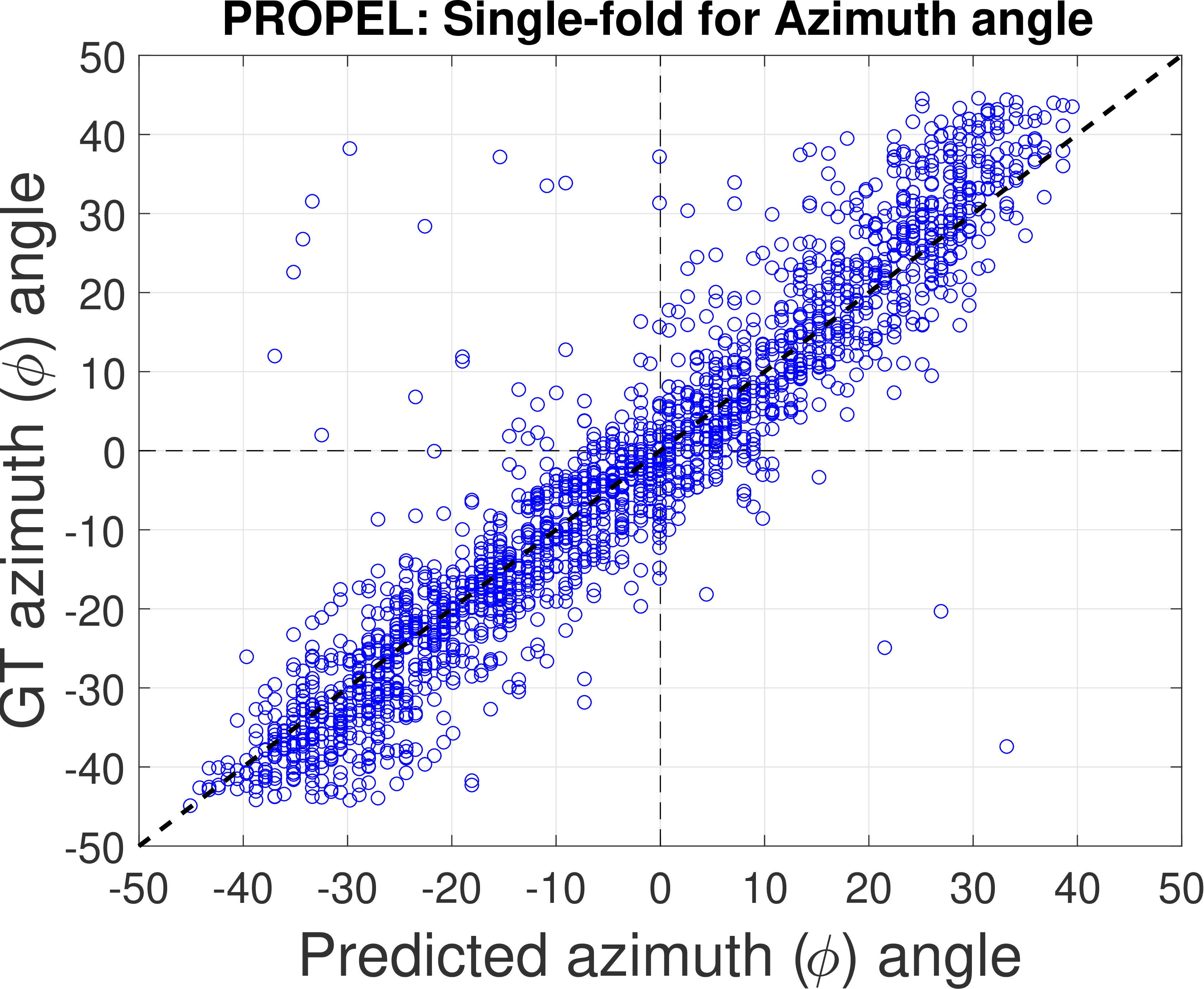}
		\includegraphics[width=0.45\linewidth, trim={0, 0, 1.5cm, 0}]{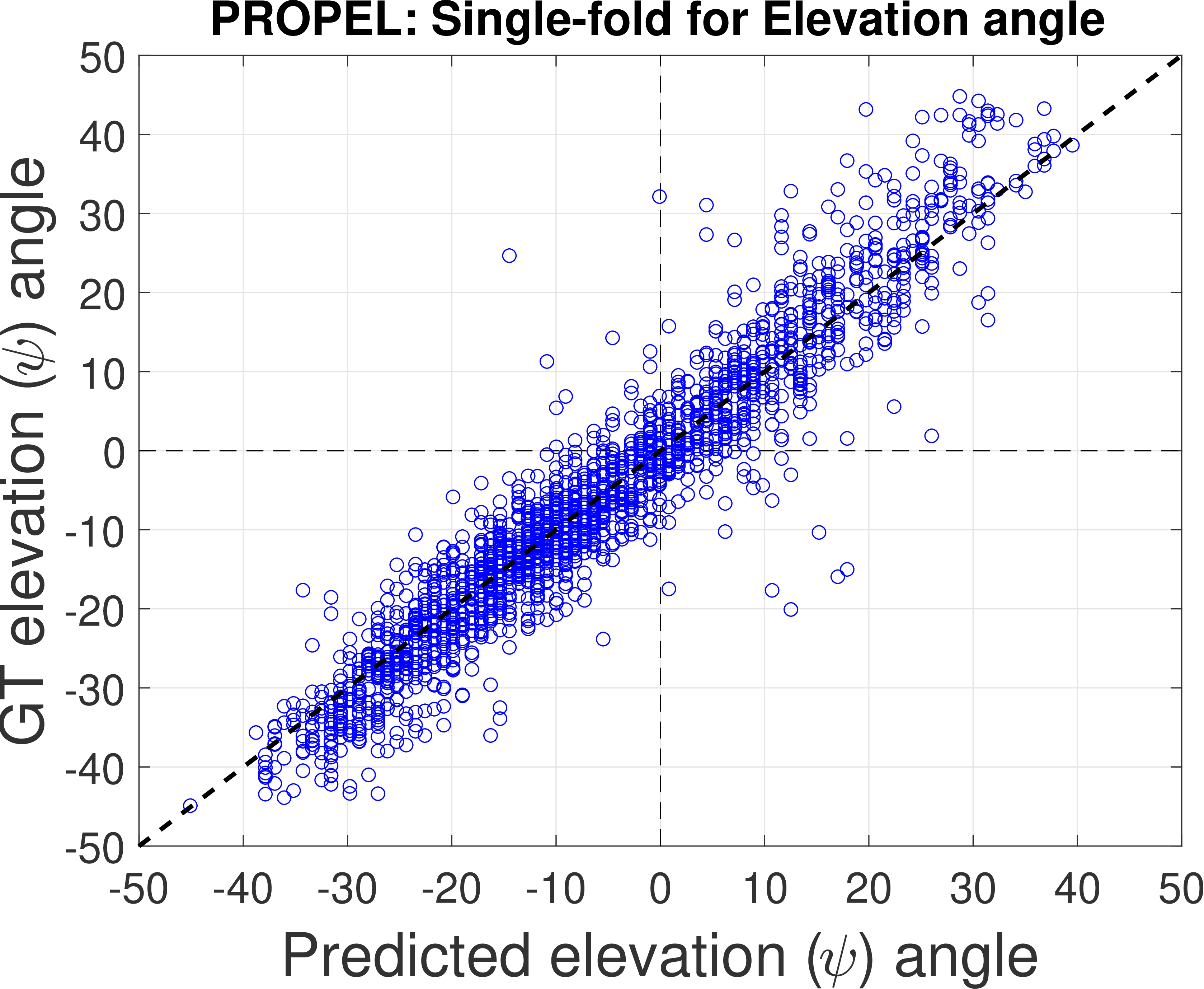}
		\par\medskip
	\end{subfigure}
	\begin{subfigure}{0.5\textwidth}
		\centering
		\includegraphics[width=0.45\linewidth, trim={1.5cm, 0, 0, 0}]{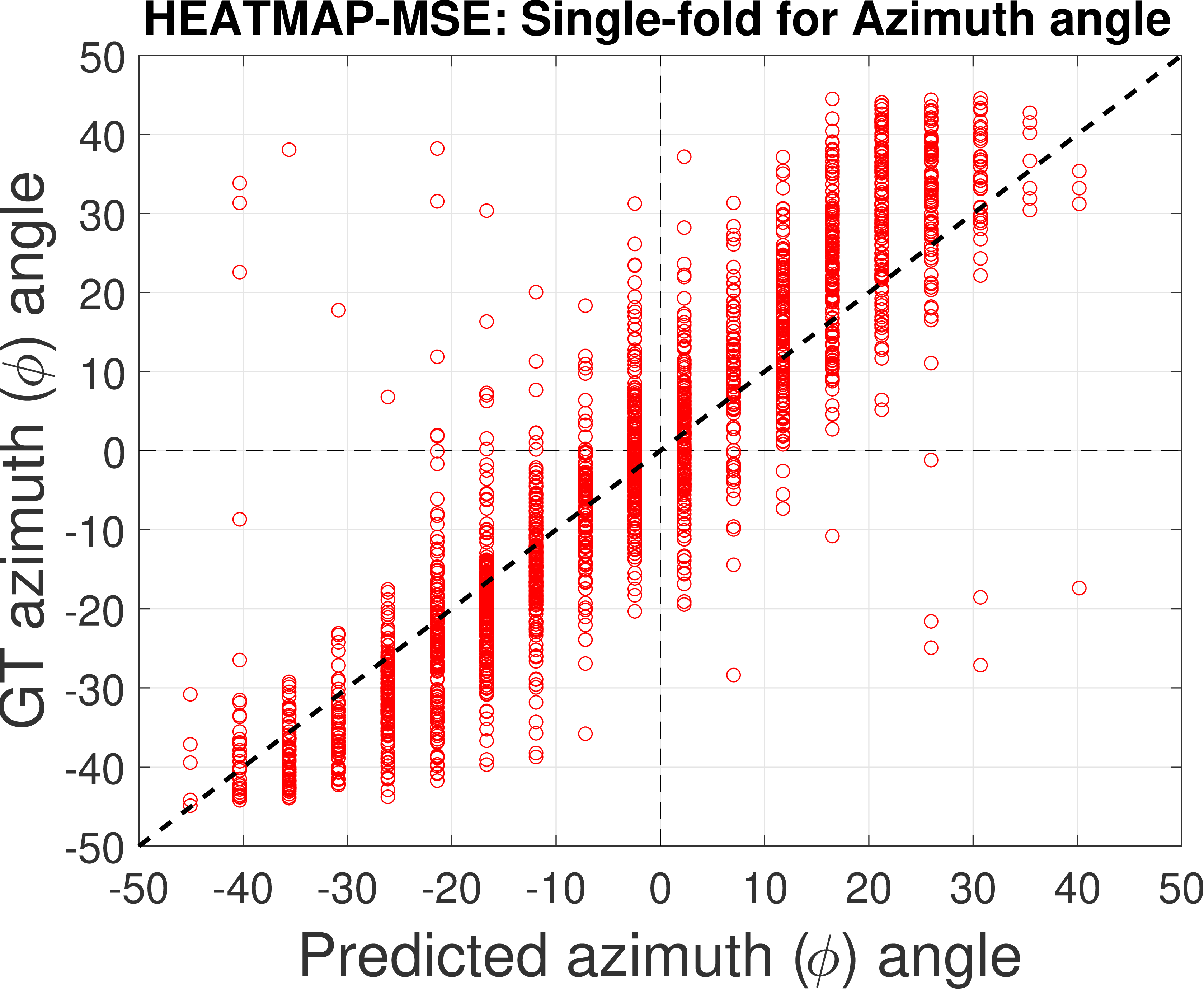}
		\includegraphics[width=0.45\linewidth, trim={0, 0, 1.5cm, 0}]{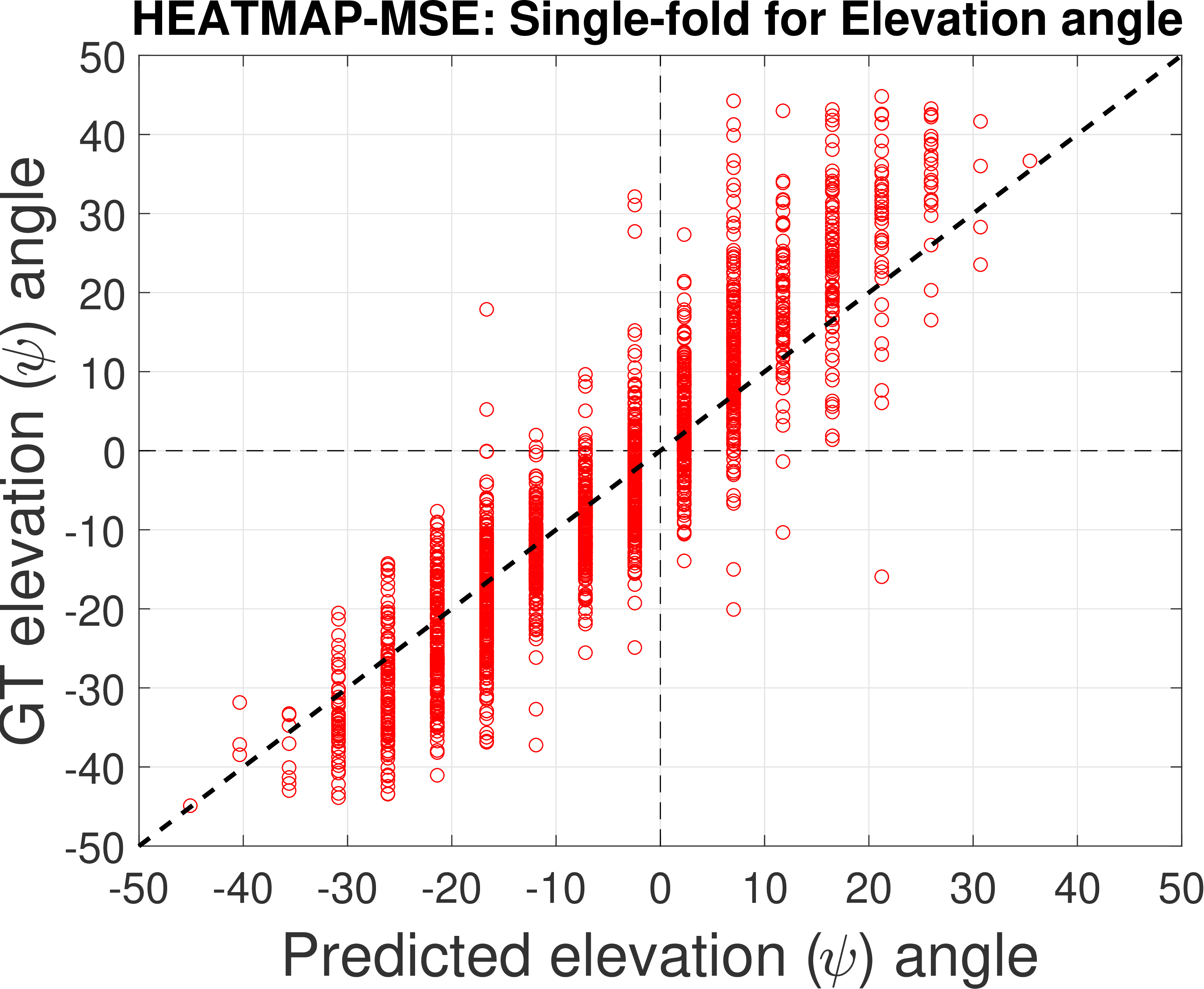}
		\par\medskip
	\end{subfigure}
	\caption{GT versus predicted angle plot using PROPEL (in blue $\textcolor{blue}{\boldsymbol{\circ}}$) and HEATMAP-MSE (in red $\textcolor{red}{\boldsymbol{\circ}}$) shows Azimuth ($\phi$) angles (left column) and Elevation ($\psi$) angles (right column). A good regressor predicts angles close to GT angles, resulting in a diagonal line.}
	\label{fig:GTVsPredPROPEL}
\end{figure}

MSE and HEATMAP-MSE are independently trained for experimental validation. Both PROPEL and HEATMAP-MSE require additional ground truth (GT) variances to be defined in Equation \ref{eq:GTDistribution}. A large variance can over-smooth the GT PDF, producing underfitted prediction PDFs. However,  a small variance yields a GT PDF that captures unwanted variations, resulting in overfitted model PDFs. We empirically found that a variance of  $(9\si{\degree})^2$, for all angles, can optimally capture the variations within the datasets, while enabling both PROPEL and HEATMAP-MSE to generalize well. Additionally PROPEL also requires the number of Gaussians $I$ in the model PDF $P_{m}$. We found that $I=2$ results in outperforming existing methods while showing the significance of having multiple components in $P_{m}$. 

Following the approach in \cite{asad2017spore}, we utilize Mean Absolute Error (MAE) and Combined Mean Absolute Error (CMAE) for evaluating the overall performance of all comparison methods.
\begin{figure*}[!tb]
	\centering
	\begin{subfigure}{.32\textwidth}
		\centering
		\includegraphics[align=c,width=0.80\linewidth, trim={1.5cm, 0cm, 0cm, 0cm}]{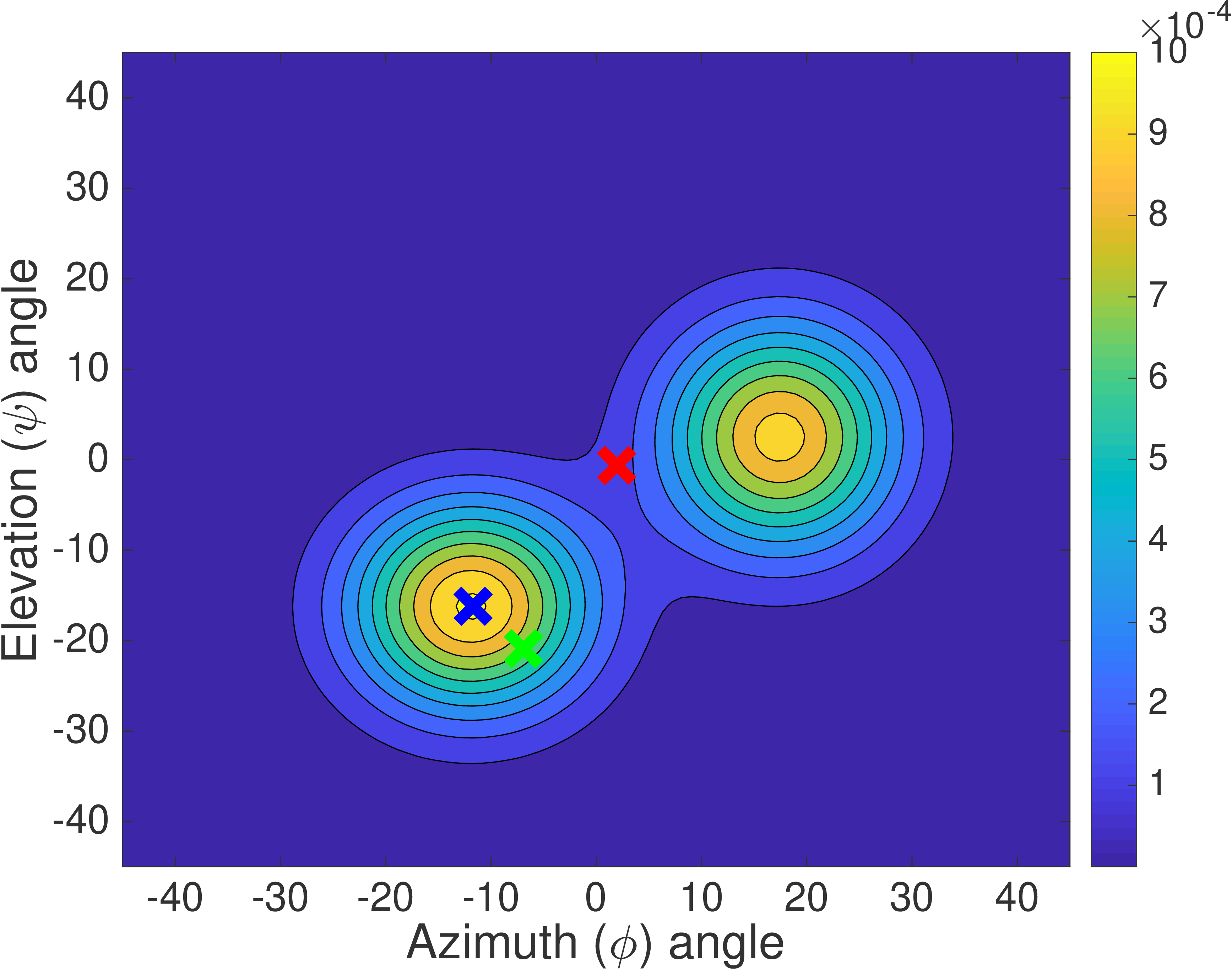}
		\includegraphics[align=c,width=0.45\linewidth]{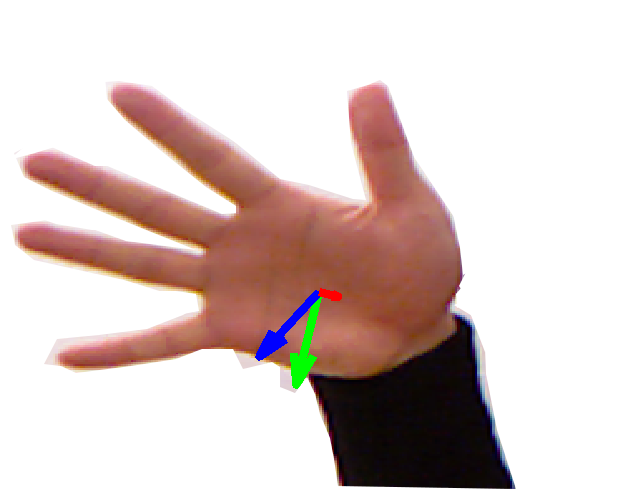}
		\caption{}
	\end{subfigure}
	\begin{subfigure}{.32\textwidth}
		\centering
		\includegraphics[align=c,width=0.80\linewidth, trim={1.5cm, 0cm, 0cm, 0cm}]{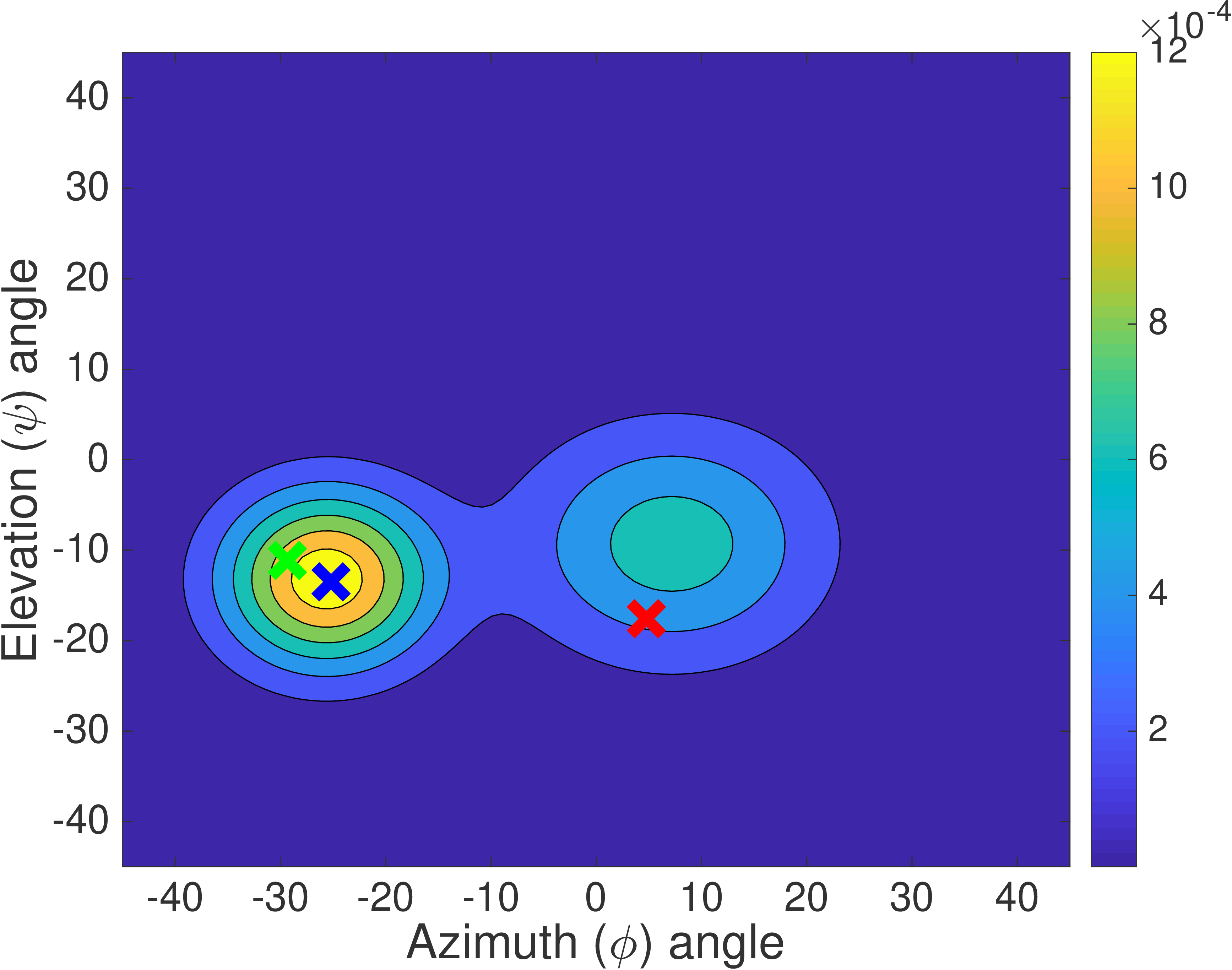}
		\includegraphics[align=c,width=0.45\linewidth]{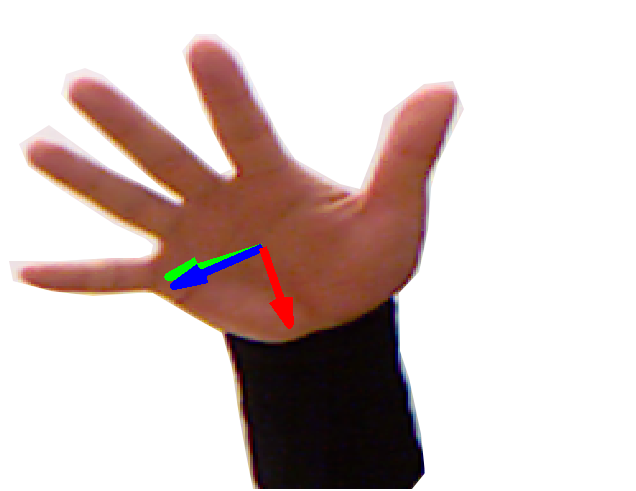}
		\caption{}
	\end{subfigure}
	\begin{subfigure}{.32\textwidth}
		\centering
		\includegraphics[align=c,width=0.80\linewidth, trim={1.5cm, 0cm, 0cm, 0cm}]{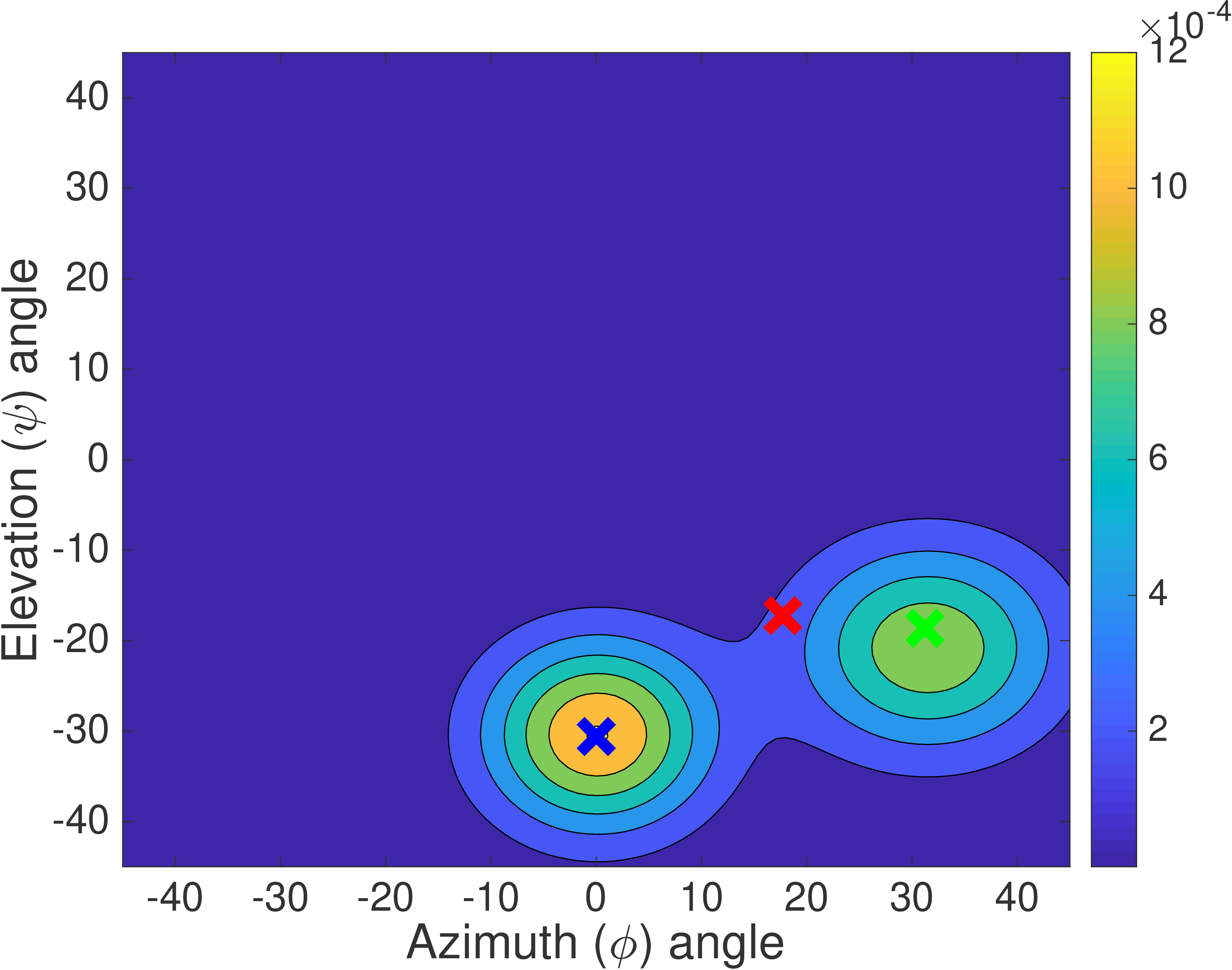}
		\includegraphics[align=c,width=0.45\linewidth]{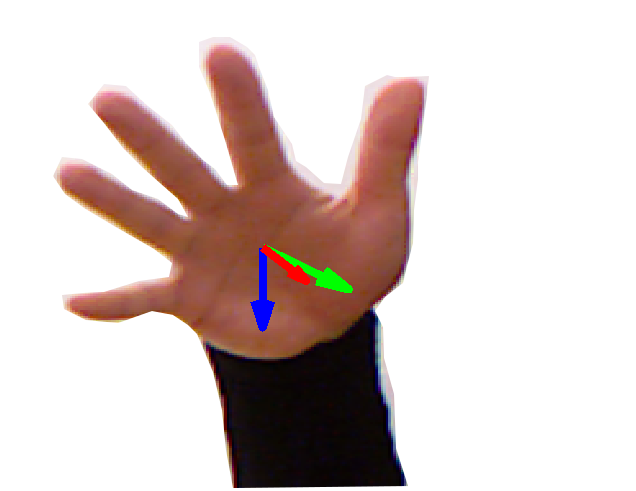}
		\caption{}
	\end{subfigure}
	
	\caption{Ability of PROPEL to handle ambiguity in predictions, showing input hand images and the corresponding predicted PDFs from PROPEL. Hand orientation angles from GT (in green $\textcolor{green}{\pmb{\uparrow}\pmb{\times}}$), and predictions from PROPEL (in blue $\textcolor{blue}{\pmb{\uparrow}}\textcolor{blue}{\pmb{\times}}$) and MSE (in red $\textcolor{red}{\pmb{\uparrow}}\textcolor{red}{\pmb{\times}}$) are shown. (a) - (b) show cases where PROPEL's PDFs successfully handles ambiguities, whereas (c) shows a failure case where PROPEL prediction contains a large error. Note that even in failure mode, PROPEL's PDF contains useful information regarding the correct prediction.}
	\label{fig:PropelVsEuc}
\end{figure*}

\subsection{Experimental Validation for Hand Orientation Regression}
We perform single-fold experimental evaluation by randomly dividing the dataset into training ($70\%$), testing ($20\%$) and validation ($10\%$) sets. This evaluates the generalization performance of the comparison methods against a scenario where the training dataset is large enough to cover variations in hand shape, color, size and orientations. Additionally, we also evaluate the performance of our method using 5-fold cross-validation, where the folds are created by grouping multiple participants' data.

Table \ref{tb:evaluationSingleFold} shows Mean Absolute Error (MAE) along with the number of model parameters for CNN methods in the single-fold validation. Further, in Fig.~\ref{fig:accuracyvsparameters} (a) we show the accuracy and model parameters trade-off. We observe that the proposed PROPEL method outperforms existing state-of-the-art in both hand-crafted feature-based as well as CNN-based hand orientation regression. Probabilistic regression in both PROPEL and HEATMAP-MSE outperforms the widely used MSE loss. This is due to the ability of probabilistic methods to provide better generalization while addressing ambiguities in cases where multiple hand orientations can result in similar projected hand shape \cite{asad2017spore}. From Fig.~\ref{fig:accuracyvsparameters} (a), we note that PROPEL requires 10x fewer model parameters as compared to HEATMAP-MSE as it enables a CNN to directly learn parameters for model PDF $P_m$. Furthermore, we note from Table \ref{tb:evaluationSingleFold} that Azimuth ($\phi$) angles have greater MAE as compared to Elevation ($\psi$) angles for all comparison methods. This is due to the variations in inter-finger separation and styles for performing pronation/supination of the forearm across different participants in the hand orientation dataset. Fig.~\ref{fig:PercentageError} provides further insight by showing the percentage of data that lies under a given error threshold for all comparison methods. It can be seen that with PROPEL $95\%$ of the data lies below $10\si{\degree}$ error, while HEATMAP-MSE has an error of $15\si{\degree}$ for the same percentage data.

In Fig.~\ref{fig:PropelVsEuc}, we show some of the ambiguous cases and compare the probabilistic output from PROPEL against predictions from MSE. PROPEL has the ability to learn multiple hypotheses, due to the presence of multiple parameterized Gaussians in the model PDF $P_m$. During training, the error is minimized for ambiguous targets where these Gaussians are able to model multiple hypotheses instead of being forced to a single hypothesis (as in MSE). On the other hand, MSE prediction \textit{tries} to fit into the target space (see Fig~\ref{fig:PropelVsEuc} (a)-(b)). Fig.~\ref{fig:PropelVsEuc} (c) shows a failure case where PROPEL prediction results in a large error. We observe that even in such a case, the predicted model PDF $P_m$ contains useful information regarding the correct prediction.

The most closely related method to PROPEL is HEATMAP-MSE as it also provides probabilistic regression by learning probability heatmaps. Table \ref{tb:evaluationSingleFold},  Fig.~\ref{fig:accuracyvsparameters} (a) and Fig.~\ref{fig:GTVsPredPROPEL} show a comparison of these methods. We make two observations from this comparison. First, PROPEL uses approximately $10\times$ fewer model parameters as compared to HEATMAP-MSE, while providing significantly better accuracy. Secondly, from Fig.~\ref{fig:GTVsPredPROPEL} (second row), quantization due to $20 \times 20$ heatmaps becomes the main source of errors in HEATMAP-MSE. Furthermore, it should be noted that with higher dimensional targets HEATMAP-MSE will require an exponentially larger number of model parameters, whereas our proposed PROPEL method will only result in increasing $n$ which is linearly related by $d = 2 \times n \times I$ to the model parameters. 

The 5-fold cross-validation helps in understanding the generalization capability of each comparison method for inferring hand orientation for unseen participants' hand shape, color and size. Table \ref{tb:evaluationSingleFold} shows results from this experiment. Once again, PROPEL outperforms all comparison methods by achieving lowest MAE, which shows that utilizing PROPEL results in a CNN model that is able to generalize better than the existing state-of-the-art for CNN regression loss.
\begin{table}[t!]
	\centering
	\setlength{\tabcolsep}{0.6\tabcolsep}
	\caption{MAE for experimental validation on BIWI dataset (*indicates the use of evaluation protocol from \cite{lathuiliere2017deep})}
	\resizebox{\columnwidth}{!}{%
		\begin{tabular}{  >{\centering\arraybackslash}m{3.2cm} >{\centering\arraybackslash}m{1.3cm}   >{\centering\arraybackslash}m{1.3cm} >{\centering\arraybackslash}m{1.3cm} >{\centering\arraybackslash}m{1.3cm} >{\centering\arraybackslash}m{1.9cm}   }
			\toprule
			\textbf{Method} & \textbf{Pitch} & \textbf{Yaw} &  \textbf{Roll} & \textbf{CMAE} & \textbf{No. of Parameters}\\ 
			\midrule
			\multicolumn{6}{c}{\textbf{Unseen Faces (training=21 users and testing=3 users)*}} \\
			\midrule
			PROPEL (proposed) & $\mathbf{3.44}\si{\degree}$ & $4.02\si{\degree}$ & $3.28\si{\degree}$ & $\mathbf{3.58\si{\degree}}$ & $50,294$\\
			MSE & $9.20\si{\degree}$ & $6.30\si{\degree}$ & $5.15\si{\degree}$ & $6.88\si{\degree}$  & $\mathbf{49,835}$\\
			Lathuili{\`e}re et al. \cite{lathuiliere2017deep} & $4.68\si{\degree} $ & $\mathbf{3.12}\si{\degree}  $ & $\mathbf{3.07}\si{\degree}$ & $3.62\si{\degree}$ & $135,847,232$\\
			Liu et al. \cite{liu20163d}  & $6.10\si{\degree} $ & $6.00\si{\degree}  $ & $5.94\si{\degree}$ & $6.01\si{\degree}$ & $595,573$\\
			Drouard et al. \cite{drouard2015head}  & $5.43\si{\degree} $ & $4.24\si{\degree}  $ & $4.13\si{\degree}$ & $4.60\si{\degree}$ & -\\
			\midrule
			\multicolumn{6}{c}{\textbf{Unseen Faces (Leave-one-out validation)}} \\
			\midrule
			PROPEL (proposed) & $5.93\si{\degree}$ & $5.81\si{\degree} $ & $\mathbf{4.53}\si{\degree}$ & $5.42\si{\degree}$ & $50,294$\\
			MSE & $7.70\si{\degree} $ & $6.70\si{\degree}$ & $5.81\si{\degree}$ & $6.74\si{\degree}$ & $\mathbf{49,835}$\\
			Liu et al. \cite{liu20163d}  & ${6.00}\si{\degree}$ & ${6.10}\si{\degree}$ & $5.70\si{\degree}$ & $\mathbf{5.17\si{\degree}}$ & $595,573$\\
			Drouard et al. \cite{drouard2015head}  & $\mathbf{5.90}\si{\degree}$ & $\mathbf{4.90}\si{\degree}$ & $4.70\si{\degree}$ & $5.93\si{\degree}$ & - \\			
			\bottomrule
	\end{tabular}}
	\label{tb:BIWIResults}
\end{table}

\subsection{Experimental Validation for Head Orientation Regression}
For this comparison, we use the established experimental protocols from
\cite{lathuiliere2017deep} and leave-one-out validation
\cite{drouard2015head}. Table \ref{tb:BIWIResults} shows comparison of PROPEL with existing state-of-the-art methods. Note that PROPEL is trained without any data augmentation, and the model is learned end-to-end unlike \cite{lathuiliere2017deep} that uses a pre-trained CNN. In contrast, PROPEL can learn to infer probabilities for any higher dimensional target. PROPEL achieves competitive results with other methods and outperforms all techniques (including the state-of-the-art) in one of three angles, in both evaluation protocols. These experiments also demonstrate the generalization of PROPEL to higher dimensional problems, for which existing heatmap-based probabilistic regression methods require exponentially more model parameters. We further visualize accuracy and model parameter trade-off in Fig.~\ref{fig:accuracyvsparameters} (b) and note that PROPEL significantly reduces the number of required model parameters, while achieving similar accuracy as the state-of-the-art method from \cite{lathuiliere2017deep}. These results indicate the potential for models trained with PROPEL, which may need much less parameters to achieve close to state-of-the-art accuracy.

\section{Limitations and Future Work}
PROPEL represents a probability distribution as a superposition of a finite number of Gaussians.  In this paper, two Gaussians are used, based on prior knowledge that there may be two possible solutions to hand or head orientation as a result of the symmetry problem.  Using a finite number of Gaussians is a strength in this paper as doing so constrains and regularises the solution space; however, for other problems which require more complex target distributions, we note a higher (possibly infinite) number of Gaussians may be required.

There are quite a few future directions for this work that can be explored. We plan to explore application of PROPEL to higher dimensional regression problems such as joint location in the hand or human body where having unconstrained probabilistic regression can improve the state-of-the-art. Another interesting aspect of PROPEL that can be studied is to include additional covariance terms in the model distributions to exploit dependence within target labels.

\section{Conclusion}
We proposed a novel PRObabilistic Parametric rEgression Loss (PROPEL) which enabled learning parametric probabilities for addressing regression problems using CNN. PROPEL is fully differentiable with an analytic closed-form solution to integrals, which allows it to be used with existing CNN architectures. A complete generalizable derivation for different level of complexity of prediction probabilities and multi-variate target dimensions was provided using Mixture of Gaussians. Comprehensive experimental validation showed that PROPEL outperforms previous state-of-the-art. It shows better generalization capabilities while reducing the number of model parameters by $10\times$. We also showed the usefulness of parametric probabilities for ambiguous cases where PROPEL handled predictions by providing multiple hypotheses. Our contribution can enable CNN regression models to have better prediction capabilities for addressing a range of problems.

\bibliographystyle{IEEEtran}
\bibliography{references}

\includepdf[pages={1}]{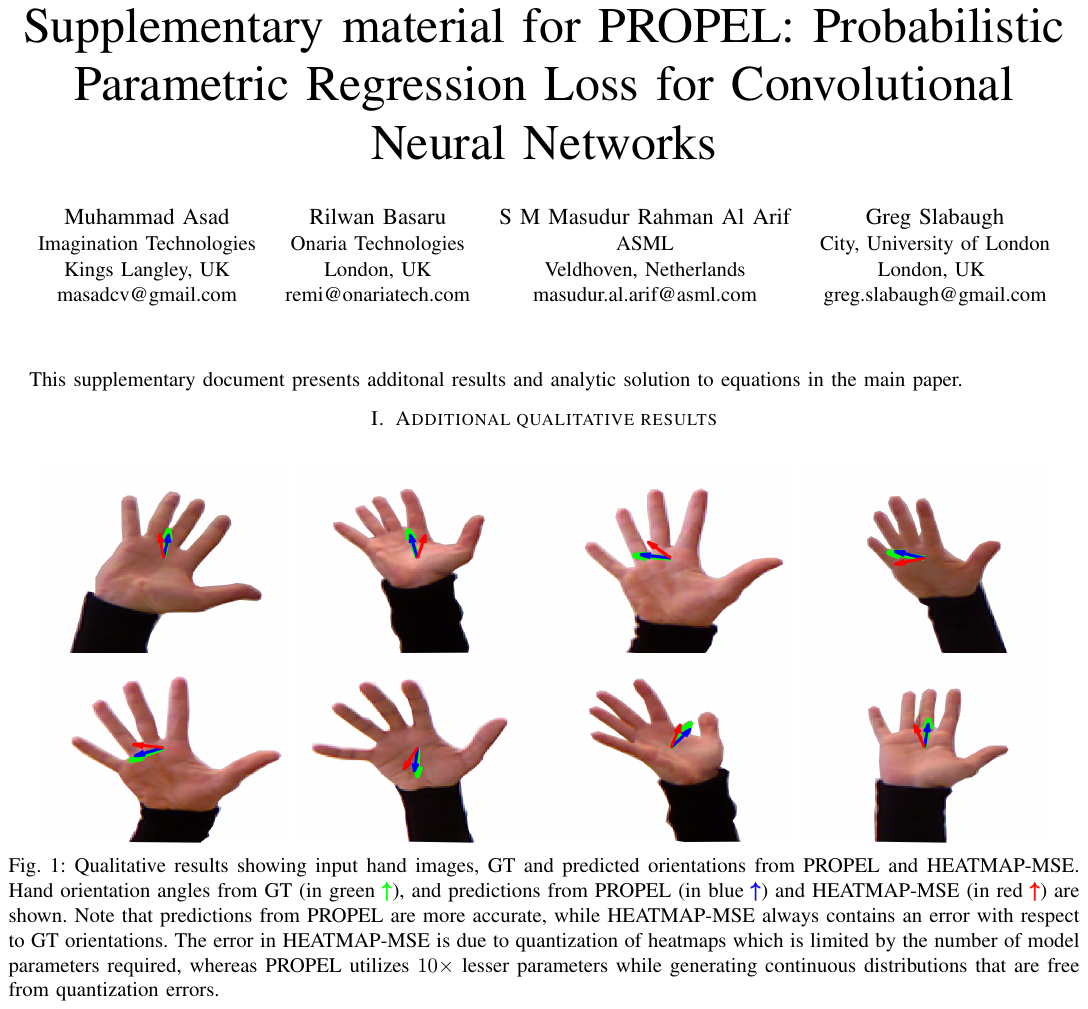} 
\includepdf[pages={2}]{supplementary/PROPEL_ICPR2020_SUP.pdf} 
\includepdf[pages={3}]{supplementary/PROPEL_ICPR2020_SUP.pdf}  

\end{document}